%% file: main.tex
\documentclass{article}

\usepackage{microtype}
\usepackage{graphicx}
\usepackage{subfigure}
\usepackage{booktabs} 
\usepackage{hyperref}

\usepackage[preprint]{icml2026}

\usepackage{amsmath}
\usepackage{amssymb}
\usepackage{mathtools}
\usepackage{amsthm}

\usepackage[capitalize]{cleveref}

\input{imports/macros}

\input{imports/packages}

\theoremstyle{plain}

\theoremstyle{definition}

\theoremstyle{remark}

\usepackage[disable,textsize=tiny]{todonotes}

\icmltitlerunning{Generalized Discrete Diffusion from Snapshots}

\begin{document}

\twocolumn[
\icmltitle{Generalized Discrete Diffusion from Snapshots}

\icmlsetsymbol{equal}{*}

\begin{icmlauthorlist}
\icmlauthor{Oussama Zekri}{ensae}
\icmlauthor{Théo Uscidda}{ensae}
\icmlauthor{Nicolas Boull\'e}{imperial}
\icmlauthor{Anna Korba}{ensae}
\end{icmlauthorlist}

\icmlaffiliation{ensae}{CREST, ENSAE, 
Institut Polytechnique de Paris, Paris, France}
\icmlaffiliation{imperial}{Department of Mathematics, Imperial College London, London, SW7 2AZ, UK}

\icmlcorrespondingauthor{Oussama Zekri}{oussama.zekri@ensae.fr}

\icmlkeywords{Machine Learning, ICML}

\vskip 0.3in
]

\printAffiliationsAndNotice{}  
\begin{abstract}
We introduce Generalized Discrete Diffusion from Snapshots ($\GDDS$), a unified framework for discrete diffusion modeling that supports arbitrary noising processes over large discrete state spaces. Our formulation encompasses all existing discrete diffusion approaches, while allowing significantly greater flexibility in the choice of corruption dynamics. The forward noising process relies on uniformization and enables fast arbitrary corruption. For the reverse process, we derive a simple evidence lower bound (ELBO) based on snapshot latents, instead of the entire noising path, that allows efficient training of standard generative modeling architectures with clear probabilistic interpretation. Our experiments on large-vocabulary discrete generation tasks suggest that the proposed framework outperforms existing discrete diffusion methods in terms of training efficiency and generation quality, and beats autoregressive models for the first time at this scale. We provide
the code along with a blog post on the project page : \href{https://oussamazekri.fr/gdds}{https://oussamazekri.fr/gdds}.
\end{abstract}

\section{Introduction}
Diffusion models~\citep{ho2020denoising,song2020score} recently became a core component of generative modeling and achieved remarkable success in high-dimensional tasks defined on continuous domains, such as image \citep{rombach2022high,saharia2022photorealistic}, audio \citep{kong2020diffwave,liu2023audioldm}, and video generation~\citep{brooks2024video,wiedemer2025video}. The extension of diffusion modeling to discrete data is of great interest since many data structures (including text, graphs, and molecules) are inherently discrete. This has led to the emergence of diffusion Large Language Models (dLLMs) \citep{lou2023discrete,li2025survey}. dLLMs offer a competitive alternative to the auto-regressive (AR) paradigm dominating language modeling~\citep{touvron2023llama,team2023gemini,liu2024deepseek} due to their ability to generate all tokens simultaneously. 

\begin{figure}[t]
    \centering
    \includegraphics[width=0.92\linewidth]{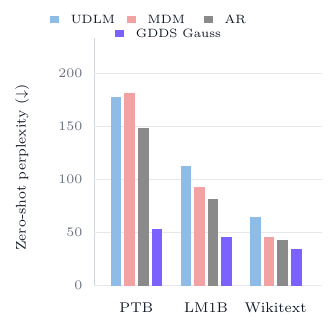}
    \caption{\textbf{Zero-shot transfer of OWT-trained models.} Zero-shot perplexity ($\downarrow$) on three representative downstream validation sets from \cref{tbl:owt_zero_shot}: PTB, LM1B, and Wikitext. Across this high-to-low perplexity range, $\GDDS$ Gauss consistently achieves the lowest transfer perplexity, highlighting the stronger generalization capability induced by semantically structured noising processes.}
    \label{fig:intro_barplot}
\end{figure}
Discrete diffusion models come in several variants, mainly differing in the choice of the noising process and how denoising is performed. 
Masked diffusion models (MDM)~\cite{sahoo2024simple,shi2024simplified,ou2024your,nie2025large} rely on a noising process where tokens are progressively replaced by a special \texttt{[MASK]} token. For uniform-state
diffusion models (USDMs)~\cite{austin2021structured,schiff2024simple,sahoo2025diffusion}, they are replaced with samples from the uniform distribution over the set of all possible tokens. These forward dynamics directly shape the reverse generation process: USDMs allow tokens to be updated continuously, whereas MDMs fix them once they are unmasked.
\begin{figure*}[hbtp]
    \centering
    \includegraphics[width=\textwidth]{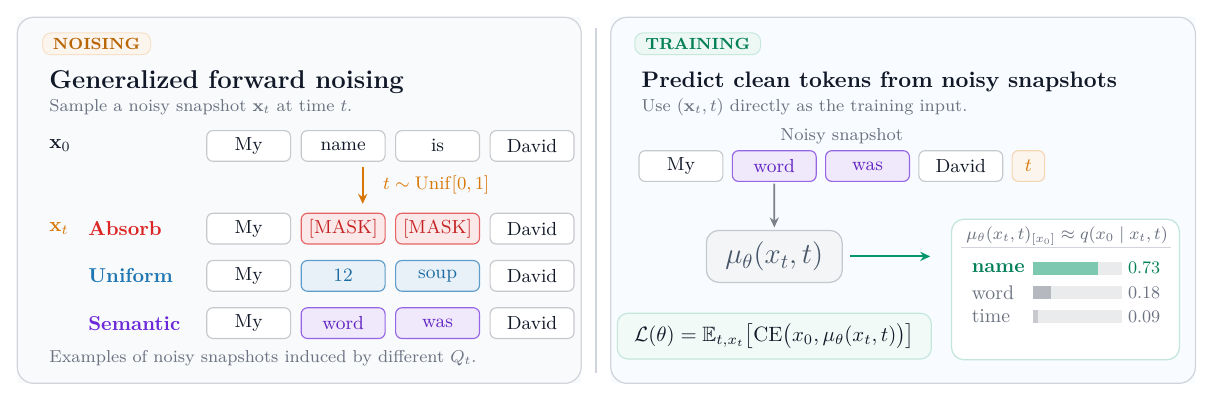}
    \caption{\textbf{Overview of GDDS.} A clean sequence $\mathbf{x}_0$ is first noised exactly by the forward CTMC at a sampled time $t\in[0,1]$, yielding a snapshot sequence $(\mathbf{x}_t,t)$. The mean parametrization is then used as a denoiser: given the snapshot, the model predicts the clean-token posterior directly from $(\mathbf{x}_t,t)$, so training is performed on snapshots rather than through a full path-wise objective.}
    \label{fig:gdds_overview}
\end{figure*}
The design space for discrete diffusion models remains surprisingly narrow. Most existing dLLMs pair a simplistic token-wise corruption rule (masking or uniform replacement) with the mean parametrization~\citep{austin2021structured}. Here, a denoiser predicts a distribution over the clean token and reverse transition probabilities are derived from this prediction through an ELBO objective. This leads to two bottlenecks: (i) the forward process is blind to any notion of neighborhood in discrete spaces (e.g., semantic proximity in language), and (ii) mean parametrization tightly constrains how denoising uncertainty can be translated into reverse dynamics, becoming increasingly restrictive beyond uniform/masked noise and at LLM scale. Advancing dLLMs calls for structure-aware noising and more flexible parametrizations that remain computationally scalable for large vocabularies and long contexts.

In this work, we generalize discrete diffusion methods by considering arbitrary noising processes and propose a tractable associated training method. We introduce the \textbf{Generalized Discrete Diffusion from Snapshots ($\GDDS$)} framework, which builds upon the most general formulation of interpolating discrete diffusion and extends it far beyond 
the restricted subclasses explored in prior work \citep{sahoo2024simple,shi2024simplified,ou2024your,von2025generalized,zhou2025next,amin2025masking}. $\GDDS$ introduces three key advances for discrete diffusion: 

\begin{enumerate}[leftmargin=*,label={\arabic*)},topsep=0pt, itemsep=0pt]
    \item \textbf{Generalized interpolating discrete diffusion:} a     mathematical framework covering arbitrary Markovian noising processes,
    encompassing all existing approaches.
    \item \textbf{Efficient noising process:} a fast forward arbitrary  corruption method for large vocabularies, requiring only column access to the rate matrix characterizing the noising process.
    \item \textbf{Parametrization and ELBO:} a principled parametrization for reverse transition probabilities, yielding a simple ELBO training objective based on snapshot samples.
\end{enumerate}
These components form the first discrete diffusion framework that is \emph{fully general} and \emph{computationally efficient}. Our experiments on large-scale language modeling tasks demonstrate \textbf{state-of-the-art modeling and generation quality}.  \Cref{fig:gdds_overview} summarizes the two ingredients behind $\GDDS$: exact forward noising to a snapshot and snapshot-level denoising.

\section{Background and Preliminaries}\label{sec:background}
This section provides background material on discrete diffusion, including the definition of rate matrices that characterize the evolution of continuous-time Markov chains, as well as common choices used in the literature.

\subsection{Discrete Diffusion}\label{sec:discrete_diff}
In discrete diffusion, the dynamics of a single token are described by a continuous-time Markov chain (CTMC) \citep{campbell2022continuous,lou2023discrete}, which is  a stochastic process $(x_t)_{t\in[0,T]}$ operating on a finite vocabulary $\cV = \{v_1,\hdots,v_m\}$. We denote by $\delta_{[x]}$ the one-hot encoding of $x\in \cV$. For a column-stochastic matrix\footnote{A column stochastic matrix is a matrix whose columns are probability distributions, hence $\sum_{i=1}^m F(i,j)=1$ with $F(i,j)\in[0,1]$ for any column $1\leq j\leq m$.} $F\in\R^{m\times m}$, we use the shorthand $F(\cdot, x) \equiv F\delta_{[x]}$ to denote the probability vector corresponding to the column indexed by $x$.

\paragraph{Forward and reverse evolution.}  Let $T>0$ be a fixed time horizon. At any time $t\in[0,T]$, the distribution of $x_t \in \cV$ is denoted by $q_t \in \Delta_m$, where $\Delta_m$ is the probability simplex over $\cV$. We set $q_0 \deq q_{\mathrm{data}}$ and $q_T \deq q_{\mathrm{ref}}$ a simple reference distribution. We represent the forward noising process through a family of Markov transition matrices $(K_t)_{t\in[0,T]}$ acting on marginals as $q_t = K_t q_0$, with $x_t \sim q_t$. Here, $K_t\in\bR^{m\times m}$ describes how probability mass flows across tokens as corruption increases. Existing discrete diffusion schemes, such as uniform or masking corruption, correspond to particular choices of $K_t$. Given a clean token $x_0 \sim q_{\mathrm{data}}$, the noised token $x_t \in \cV$ at time $t$ is drawn from the categorical distribution:
\begin{equation}\label{eq:main_marginal}
    q_t(x_t\mid x_0) = \Cat(x_t;K_t(\cdot,x_0))=K_t(x_t,x_0).
\end{equation}
 
While $K_t$ can be specified directly, we focus on the principled setting where these  noising operators are induced by a continuous-time Markov process with (possibly time-inhomogeneous)  rate matrix (also called infinitesimal generator) $Q_t$. This matrix is defined as 
$Q_t(i,j) = \lim_{h \downarrow 0} \frac{\mathbb{P}(x_{t+h}=i|x_t=j) - \delta_{ij}}{h}$ for $i\neq j$, and diagonal entries enforcing conservation of mass $Q_t(j,j)=-\sum_{i\neq j}Q_t(i,j)$. In this case, $K_t$ is defined as the solution to the Kolmogorov forward equation:
\begin{equation}\label{eq:kolmogorov_forward}
    \frac{\ud K_t}{\ud t} = Q_tK_t,\quad K_0=I_m, \quad \text{for } t\in[0,\horizon].
\end{equation}
Solving \cref{eq:kolmogorov_forward} yields
$K_t = \cT\exp(\int_0^t Q_s\ud s),$ where $\cT$ is the time-ordering operator (see~\cref{app:exp}). Since each column $K_t(\cdot, x_0)$ for $x_0 \in \mathcal{V}$ lies in the probability simplex $\Delta_m$ and corresponds to the forward marginal $q_t(\cdot \mid x_0)$ as given in \cref{eq:main_marginal}, a noisy token $x_t$ can be obtained by directly sampling from this categorical distribution, without simulating the underlying continuous-time trajectory.
Moreover, since $q_t = K_t q_0$, token distributions evolve according to the Kolmogorov forward
equation for marginals:
\begin{equation}\label{eq:fokker_planck}
\frac{\ud q_t}{\ud t} = Q_tq_t, \quad x_t\sim q_t,\quad\text{for } t\in[0,\horizon].\end{equation}
The time reversal of this equation \citep{kelly2011reversibility} defined through $p_t := q_{\horizon-t}$ is
\begin{equation}\label{eq:fokker_planck_timerev}
\frac{\ud p_{t}}{\ud t} = \overline{Q}_{t} p_{t},\quad x_{T-t}\sim p_t, \quad\text{for }t\in[0,\horizon],
\end{equation}
where $\overline{Q}_{t}(i, j) = (q_t(i)/q_t(j)) Q_{t}(j, i)$ if $i \neq j$ and $\overline{Q}_{t}(i, i)=-\sum_{j \neq i} \overline{Q}_{t}(j,i)$. Without loss of generality, we select $\horizon = 1$ as any bounded interval $[0,\horizon]$ can be rescaled to $[0,1]$ through the change of variable $t \mapsto t/\horizon$. In discrete diffusion, a neural network learns to simulate the reverse dynamics given by \cref{eq:fokker_planck_timerev} to reconstruct a clean data $x_0$ from a fully noised quantity $x_1$.

\paragraph{Rate matrices.}

At any time $t\in[0,1]$, the forward and reverse CTMCs evolve according to rate matrices $Q_t$ and $\overline{Q}_t$ with non-negative off-diagonal entries $Q_t(i,j)\geq 0$ for $i\neq j$, and diagonal entries verifying $Q_t(j,j)=-\sum_{i\neq j}Q_t(i,j)$, and analogously for $\overline{Q}_t$. We factorize them into \emph{exit rates} and \emph{jump kernels} as
\begin{align*}
Q_t &= (F_t - I_m)\,\diag(f_1(t),\dots,f_m(t)),\\
\overline{Q}_t &= (R_t - I_m)\,\diag(r_1(t),\dots,r_m(t)),
\end{align*}
where $f_j(t)=\sum_{i\neq j} Q_t(i,j)$ and $r_j(t)=\sum_{i\neq j}\overline{Q}_t(i,j)$ are the (forward and reverse) exit rates, controlling \emph{how often} the chain leaves state $j$. The matrices $F_t$ and $R_t$ specify \emph{where} the chain jumps when it leaves a state and are defined as column-stochastic matrices:
\[
F_t(i,j) = 
\begin{cases}
\dfrac{Q_t(i,j)}{f_j(t)}, & f_j(t)>0,\\
0, & f_j(t)=0,\ 
\end{cases}\text{ for } i\neq j,
\]
and $F_t(j,j) = 1 - \sum_{i\neq j} F_t(i,j)$, analogously for $R_t$. Note that this factorization is not exploited in the literature. 

\subsection{Designs of the rate matrix} \label{sec:design_matrix}
A common choice to simplify the forward noising process is to select equal forward exit rates: $f_1(t)=\ldots=f_m(t)=f(t)$, and a time-independent forward jump kernel $F_t = F$. In this case, $Q_t = f(t)(F - I_m),$ and a single matrix $F\in\bR^{m\times m}$ must be stored. In this model, the time-ordered exponential $\mathcal{T}$ simplifies to a standard matrix exponential, 
and ensures that $K_t$ admits the following closed form:
\[
K_t = \exp\left(\bar{f}(t) (F-I_m)\right),
\quad \text{where }\bar{f}(t) = \int_0^t f(s)\ud s.
\]
Being able to sample from columns of the matrix exponential $K_t$ is crucial to design a scalable noising process.

\paragraph{Usual forms of $F$.} For typical vocabulary sizes used in language models ($m= 50{,}257$ for GPT-2; \citealt{radford2019language}), storing a dense $F\in\mathbb{R}^{m\times m}$ requires more than $2.5\times10^9$ parameters ($\approx20$GB in double precision) and each matrix-vector products involving $F$ costs $\cO(m^2)$ time complexity, making it computationally impractical. Hence, forward kernels are usually highly structured \citep{austin2021structured,campbell2022continuous,lou2023discrete}, such as the ones related to the uniform and mask noising processes:
\begin{equation} \label{eq:choice_q}
    F^{\rm uniform} \coloneqq
    \frac{1}{m}\unit\unit^\top, \quad 
    F^{\rm absorb} \coloneqq
    \delta_{\texttt{[MASK]}}\unit^\top,
\end{equation}
for which $K_t$ admits closed‐form expressions \citep{austin2021structured,lou2023discrete}, enabling efficient noising. Since these $F$ matrices are idempotent, the exponential matrix writes $K_t=\exp(-\overline{f}(t))I_m + (1-\exp(-\overline{f}(t)))F$. However, these restrictive structures impose rigid corruption patterns on the tokens, motivating  our flexible approach.

\section{Forward noising with diffusion} \label{sec:forward_noise}

\subsection{Generalized interpolating discrete diffusion}\label{sec:general_interpolating}
We consider a time-differentiable, decreasing, \emph{mixing rate} $t\mapsto \alpha_t: [0,1] \to [0,1]$ such that $\alpha_0 = 1$, $\alpha_1 = 0$, and $\alpha_t<1$ for $t>0$. We introduce a time-differentiable column-stochastic mixing matrix $\Pi_t \in \mathbb{R}^{m \times m}$, which specifies how probability mass is redistributed across tokens as noise increases, along with its \emph{interpolating matrix} as\footnote{
We assume that, for every $t\in(0,1)$, $K_t$ is invertible and the solution $\smash{v^{(t,x)}}$ to the linear system $\smash{K_t^\top} \smash{v^{(t,x)}}=\smash{\dot{K}_t^\top\delta_x}
$ satisfies $\smash{v^{(t,x)}_{[y]}}\geq 0$ for $y\neq x$ to guarantee that $K_t$ induces a valid  CTMC.}
\begin{equation}\label{eq:main_interpolating_matrix}
K_t \deq \alpha_t I_m + (1 - \alpha_t)\Pi_t, \quad t\in[0,1].
\end{equation}
Here, $\Pi_t$ encodes the structure of the noising mechanism and $\alpha_t$ its intensity. This formulation recovers common discrete diffusion schemes as special cases, such as masked or uniform. Yet, more general choices of $\Pi_t$ allow for structured and token-dependent corruption mechanisms. The rate matrix $Q_t$ associated with $K_t$ is given in \cref{prop:rate_matrix}.

\begin{boxprop}\label{prop:rate_matrix}
Let $t\geq 0$ and denote by $\dot{K}_t$ the time derivative of $K_t$. The rate matrix induced by \cref{eq:main_interpolating_matrix} is $Q_t = \dot{K}_tK_t^{-1}$.
\end{boxprop}

Choosing a column-constant mixing matrix $\Pi_t = \pi_t \unit^\top$ (a rank-one form with $\pi_t\in\simplex$) in \cref{eq:main_interpolating_matrix} yields the GIDD formulation of \citep[Lem.~3.6]{von2025generalized}, namely $Q_t = \tfrac{\dot{\alpha}_t}{\alpha_t} I_m + (1-\alpha_t)\dot{\pi}_t \unit^\top - \tfrac{\dot{\alpha}_t}{\alpha_t}\pi_t \unit^\top$ (see \cref{cor:gidd}). However, this formulation encompasses all existing frameworks, including \citep{zhou2025next} and GenMD4 \citep{shi2024simplified}, unlike GIDD \citep{von2025generalized}.

\paragraph{Expressiveness.} Reversely, given any rate matrix $Q_t$, we aim to find a mixing matrix $\Pi_t$ such that the interpolating matrix $K_t$ defined by \cref{eq:main_interpolating_matrix} coincides with a solution to \cref{eq:kolmogorov_forward}; which induces the  marginal $(q_t)_{t\geq 0}$ as in \cref{eq:fokker_planck}.

\begin{boxprop}\label{prop:mixing_matrix}
    Let $\alpha_t$ a mixing rate such that $\dot{\alpha}_0 <0$ and $Q_t$ a     rate matrix. There exists a unique mixing matrix $\Pi_t\in\R^{m\times m}$ such that for all $t\in[0,1]$, 
    \[K_t = \alpha_t I_m + (1 - \alpha_t)\Pi_t= \cT\exp\left(\int_0^tQ_s\ud s\right).\]
\end{boxprop}

Following \cref{prop:mixing_matrix}, if $\Pi_t$ is known in closed-form, then simulating the noising process becomes possible. Indeed, $q_t(\cdot\mid x_0)= \Cat(\cdot;K_t(\cdot,x_0))$ requires only the evaluation of the column $\Pi_t(\cdot,x_0)$ instead of costly matrix exponentiations. While columns are known in closed form for uniform or masked schemes, this is generally not the case for an arbitrary $\Pi_t$.

\subsection{Efficient forward noising through uniformization}\label{sec:efficient_forward}

Since computing the marginals $q_t(\cdot \mid x_0)$ exactly is generally intractable, we employ an exact noising procedure based on uniformization. Classical uniformization provides an exact Poisson-based representation of the matrix exponential $K_t$~\citep{jensen1953markoff,stewart2009probability}. Here, we use the same procedure to generate exact forward samples $x_t \sim q_t(\cdot\mid x_0)$ without requiring  exact knowledge of these marginals; hence avoiding computing the exponential. Following \cref{prop:mixing_matrix}, the interpolating matrix in \cref{eq:main_interpolating_matrix} is expressive enough to represent any rate matrix $Q_t$. Recall that any such $Q_t$ can be written in factored form as $Q_t = (F_t - I_m)\diag(f_1(t),\dots,f_m(t))$. To simplify the exposition, we focus on the shared exit rates case:
\begin{equation}\label{eq:simplified_rate}
    Q_t = f(t)(F_t - I_m),
\end{equation}
which preserves the transition structure encoded in $F_t$, while making the uniformization-based noising process significantly easier to implement. This result can be extended to general non-shared exit rates through Poisson thinning. We denote by $\bar{f}(t) = \int_0^t f(s)\ud s$ the integrated exit rate and set the mixing rate to $\alpha_t = \exp(-\bar f(t))$ for $t\in[0,1]$.

\begin{boxprop}[Uniformization]\label{prop:mixing_matrix_uniformization}
    Consider a rate matrix $Q_t$ of the form \eqref{eq:simplified_rate} and the mixing rate $\alpha_t = \exp(-\bar{f}(t))$, where $\bar{f}(t) = \int_0^t f(s)\ud s$. Let $(N_t)_{t\in[0,1]}$ be a non-homogeneous Poisson process with intensity $\bar{f}(t)$, and denote by $0 < T_1 < \ldots < T_{N_t} \leq t$ its jump times on $[0,t]$. The unique matrix $\Pi_t$ provided by \cref{prop:mixing_matrix} is $\Pi_t = \bE[F_{T_{N_t}}\hdots F_{T_{1}}\mid N_t\geq 1]$.
\end{boxprop}

Therefore, computing $q_t(\cdot\mid x_0)$ at any $t\in[0,1]$ amounts to computing a column of $\bE[F_{T_{N_t}}\hdots F_{T_{1}}\mid N_t\geq 1]$, which can be done \emph{approximately} even when the vocabulary size $m$ is large~\citep{dingle2004uniformization}. If we only need to draw samples $x_t \sim q_t(\cdot\mid x_0)$ rather than evaluate the full distribution, we can instead sample \emph{exactly} by performing $N_t$ transitions with the matrix $F_t$, using only Poisson sampling and column access to $Q_t$ (see \cref{app:uniform_sampling}). This procedures implicitly builds a discrete-time Markov chain $z_k =x_{T_k}$ for all $1\leq k\leq N_t$ initialized at $z_0=x_0$. \Cref{alg:forward-tokenlevel} details the resulting token-level noising procedure and returns the noised token $z_{N_t} = x_{T_{N_t}}$, which coincides exactly with $x_t\sim q_t(\cdot\mid x_0)$ following \cref{prop:mixing_matrix_uniformization}. 
\begin{algorithm}[htbp]
\caption{Exact general noising, token level}
\label{alg:forward-tokenlevel}
\begin{algorithmic}[1]
\STATE \textbf{Input:} clean token $z_0 = x_0$, time $t \in[0,1]$, intensity $\bar{f}(t)$, rate matrix $Q_t$ as in \cref{eq:simplified_rate}
\STATE Sample number of jumps $N_t \sim \Poiss(\bar{f}(t))$
\STATE Sample and sort the jump times as $T_{1} < \ldots < T_{N_t}$
\FOR{$k = 1$ to $N_t$}
    \STATE Sample jump $z_{k} \sim F_{T_k}(\cdot , z_{k-1})$
\ENDFOR
\STATE \textbf{return} noised token $x_t = z_{N_t}$ and jumps $(z_k,T_k)_{k=1}^{N_t}$
\end{algorithmic}
\end{algorithm}

\cref{alg:forward-tokenlevel} enables efficient noising for any continuous-time noising process (beyond masked and uniform), requiring only column access to the rate matrix $Q_t$ (instead of its generally intractable matrix exponential). This procedure generalizes easily to a parallel sequence-level algorithm for a sequence $\bx_0 = x^1_0\ldots x^n_0$ of length $n\geq 1$ (see \cref{alg:forward-seqlevel} in \cref{app:uniform_sampling}). While the time input can be any value $t\in[0,1]$, selecting $t=1$ yields a full forward noising path of the form $\omega=\{N_1,(z_k,T_k)_{k=1}^{N_1}\}$.

\section{Reverse learning: aligning the generative model, and the objective}

Readers mostly interested in the implementation and the loss function may refer to \cref{sec:snapshot,alg:main_algo}.

\subsection{The core mismatch in reverse parametrization}

A common choice in the discrete diffusion litterature to simulate the reverse dynamics is to use the \emph{mean} parametrization (also known as $x_0$-parametrization). Concretely, $\mu_\theta:\cV\times[0,1]\to\Delta_m$ is a neural network outputting a probability vector on the token space $\cV$, which aims to approximate the posterior of the clean token from \textit{snapshots latents} $s=(x_t,t)$ generated by the forward noising process, i.e., $\mu_\theta(x_t,t)_{[x_0]}\approx q(x_0\mid x_t,t)$. It is often plugged into the reverse-time model via Bayes' rule as
\begin{equation}
\label{eq:param_old_full}
p^{\theta,\mathrm{path}}_{u\mid t}(x_u\mid x_t)
= \sum_{x_0\in\cV} q(x_u\mid x_t,x_0) \mu_\theta(x_0\mid x_t,t),
\end{equation}
where $q(x_u\mid x_t,x_0)=\frac{q_{t\mid u}(x_t\mid x_u) q_{u}(x_u\mid x_0)}{q_{t}(x_t\mid x_0)}$ and $q_{t\mid u}(x_t\mid x_u)$ denotes the forward conditional from time $u \in [0,1]$ to $t \in [0,1]$ with $u\leq t$. However, plugging $\mu_\theta$ into the reverse-time model through \cref{eq:param_old_full} does not generally enforce $\mu_\theta(x_t,t)_{[x_0]}\approx q(x_0\mid x_t,t)$. This construction glues the mean denoiser to the entire reverse CTMC: the same $\mu_\theta$ controls \emph{when} the chain jumps (reverse intensities) and \emph{where} it jumps (reverse destinations), creating a training burden mismatch. Our insight is that \textbf{the mean network naturally parametrizes a snapshot generative model} that should be trained to actually achieve $\mu_\theta(x_t,t)_{[x_0]}\approx q(x_0\mid x_t,t)$, whereas modeling the reverse CTMC calls for a jump network $\sj_\theta$ designed directly for the path-wise generative model with \emph{path-wise latents} $\omega$. To align the objective with the generative object, we first parametrize the reverse CTMC directly by disentangling jump times and jump destinations. Then, we focus on how one should design a snapshot generative model from the mean parametrization.

\subsection{Path-wise model and loss function} \label{sec:path_wise}

\paragraph{Jump-states parametrization.}
Inspired by the factorization $\overline{Q}_t = (R_t - I_m)\,\diag(r_1(t),\dots,r_m(t))$ of the true reverse generator, we directly learn $R_t$ while keeping the exit-rate schedule $\{r_i(t)\}_{i=1}^m$ fixed to the true reverse rates. Note that even when the forward process uses shared exit rates as in \cref{eq:simplified_rate}, the reverse exit rates are not shared in general (see e.g. the masked diffusion example in \cref{app:md4_mdlm}). We consider a neural network $\sj_\theta:\cV\times [0,1]\to \Delta_m$ that yields the following \emph{jump-states} parametrization:
\begin{equation}\label{eq:jump_parametrization}
\overline{Q}_t^\theta = (R_t^\theta - I_m)\,\diag(r_1(t),\dots,r_m(t)),
\end{equation}    
where $R_t^\theta \in \mathbb{R}^{m \times m}$ is the column-stochastic infinitesimal reverse jump kernel (\emph{where} the chain jumps) defined by $R_t^{\theta}\delta_{[x_t]}\deq \sj_\theta(x_t,t)$, and the exit rates $r_i(t)$ (\emph{when} the chain jumps) remain fixed. This parametrization of \cref{eq:fokker_planck_timerev} is fundamentally different from the score parametrization of \cite{lou2023discrete}, the schedule-conditioned parametrization of \cite{amin2025masking} and the parametrization in \cref{eq:param_old_full} (cf.~\cref{app:reverse_process}).

\paragraph{Path-wise ELBO.} We derive the Evidence Lower Bound (ELBO) associated with our jump-states parametrization given by \cref{eq:jump_parametrization} in \cref{prop:elbo}. This parametrization is \emph{key} to obtain a simple, CTMC-aligned, ELBO with a clean learning objective: a weighted cross-entropy that matches the model reverse jump kernel to the ideal reverse jump kernel, with $\theta$-independent weights given by the reverse exit rates $r_{[x_t]}(t)$. The result holds for any forward rate matrix $Q_t$, as the interpolating family $K_t = \alpha_t I_m + (1-\alpha_t)\Pi_t$ can represent an arbitrary rate matrix (\cref{eq:main_interpolating_matrix,prop:mixing_matrix}). Here, $p^{\theta,\mathrm{path}}_{t}$ is induced by \cref{eq:fokker_planck_timerev} where we replace $\smash{\overline{Q}_t}$ by $\smash{\overline{Q}_t^\theta}$.

\begin{boxprop}[Path-wise ELBO]\label{prop:elbo}
    Let $x_0\in\cV$, the ELBO is $\log p_0^{\theta,\mathrm{path}}(x_0) \geq -\cL_{x_0}^{\mathrm{path}}(\theta)+C^{\mathrm{path}}_{x_0}$, where
    \[\resizebox{\linewidth}{!}{$\displaystyle
    \cL_{x_0}^{\mathrm{path}}(\theta) = \!\int_0^1\! \bE_{x_t\sim q_t(\cdot\mid x_0)}\!\!\left[r_{[x_t]}^{x_0}(t)\CE(R_t^{x_0},R_t^\theta)\big|_{x_t}\!\right]\!\!\ud t,$}
    \]
    and $C^{\mathrm{path}}_{x_0}$ is independent of $\theta$. Here, $\CE(R_t^{x_0},R_t^\theta)\big|_{x_t}$ denotes the cross-entropy between the vectors $R_t^{x_0}(\cdot,x_t)$ and $R_t^\theta(\cdot,x_t) = \sj_\theta(x_t,t)$. $R_t^{x_0}$ and $r_{[x_t]}^{x_0}(t)$ denote respectively the true conditional reverse jump kernel and its associated exit rate (see \cref{app:def:Rx0}).
\end{boxprop}
In the masked diffusion case, where $\Pi_t=\delta_{\texttt{[MASK]}}\unit^\top$, our jump parametrization and ELBO coincide with the parametrization \eqref{eq:param_old_full} and ELBO used in prior work \citep{sahoo2024simple,shi2024simplified,ou2024your}; see \cref{app:md4_mdlm}. Beyond this setting, other objectives such as \cite{von2025generalized,zhou2025next,lou2023discrete,amin2025masking} apply to broad classes of noising processes but do not isolate such a clean weighted cross-entropy signal and involve additional terms. Indeed, we show in \cref{app:elbo} that $\cL_{x_0}^{\mathrm{path}}(\theta)$ can be written as
\[
\resizebox{\linewidth}{!}{$\displaystyle \cL_{x_0}^{\mathrm{path}}(\theta)\! = \!\!\int_0^1\!\!\! \bE_{x_t\sim q_t(\cdot\mid x_0)}\!\!\left[\!\sum_{y\neq x_t}\!\!\frac{q_t(y\mid x_0)}{q_t(x_t\mid x_0)}Q_{t}(x_t,i)(-\log \sj_\theta(x_t,t)_y)\!\right]\!\!\!\ud t.$}
\]
This loss remains useful beyond masking: when the forward marginal is tractable (typically, when $\Pi_t$ is known so that $q_t(\cdot\mid x_0)$ can be evaluated; e.g., in the masked or uniform case), $\cL_{x_0}^{\mathrm{path}}(\theta)$ is fully computable and directly trains the \emph{path-wise} reverse CTMC. However, we seek to avoid knowledge of $q_t(\cdot\mid x_0)$ for a general CTMC $Q_t$, for which the associated $\Pi_t$ is unknown.

\paragraph{Campbell estimator.}
To mitigate this issue, we introduce a path-wise \emph{Campbell estimator} that is a clean rewriting of the path-wise loss. It consists of applying Campbell's formula \citep{campbell1909study,last2018lectures} to $\smash{\cL_{x_0}^{\mathrm{path}}(\theta)}$, which transforms the integral into a sum over the whole noising path in $[0,1]$ given by the Poisson process of \cref{alg:forward-tokenlevel}. Importantly, this expression does not involve any other quantity than the network output and the uniformization path produced by \cref{alg:forward-tokenlevel}, making it computable even when $q_t(\cdot\mid x_0)$ (i.e., $\Pi_t$) is unknown.

\begin{boxprop}[Campbell estimator]\label{prop:campbell}
Let $x_0\in\cV$ and $\omega\sim q_{[0,1]}(\cdot\mid x_0)$ denote the full forward noising path produced by \cref{alg:forward-tokenlevel}. Writing $\omega = \{N_1,(z_k,T_k)_{k=0}^{N_1}\}$ for its jump counts and marks, we have
\[
\resizebox{\linewidth}{!}{$\displaystyle \cL_{x_0}^{\mathrm{path}}(\theta) = \bE_{\omega\sim q_{[0,1]}(\cdot\mid x_0)}\!\left[\sum_{k=1}^{N_1}-\log \sj_\theta(z_k,T_k)_{[z_{k-1}]}
\right]\!.$}
\]
\end{boxprop}

This loss is reminiscent of any-order AR objectives (e.g., XLNet; \citealt{yang2019xlnet}), except that each term predicts the pre-jump token $z_{k-1}$ from the post-jump noised context $(z_k,T_k)$, and the factorization order is induced by Poisson jump times (closer in spirit to denoising permutation objectives such as MPNet; \citealt{song2020mpnet}). As a result, training requires evaluating many snapshot-wise conditionals along a single path and, in practice, calls for a two-stream mechanism (separating a content stream encoding the clean tokens from a query stream used to predict the target token $[z_{k-1}]$, as in XLNet), which is not naturally aligned with standard transformer architectures and empirically underperforms them (see \cref{app:campbell_model}).

\subsection{Snapshot model and loss function}\label{sec:snapshot}

This observation motivates our approach: if powerful models mostly train on snapshot noised contexts, why should the variational latent variable be the \emph{entire path} $\omega$? Therefore, we consider a \emph{snapshot-latent} variational formulation, where $s=(x_t,t)$ replaces $\omega$ as the latent variable. Crucially, this choice also aligns with the perspective of \citet{li2025back} that denoising models should predict the \emph{clean} quantity through the mean parametrization, rather than a noised quantity as the jump states parameterizations do. This aligns the mean parametrization $\mu_\theta$ to its coherent generative model, and yields an objective that is directly compatible with standard architectures for any general noising process.

\paragraph{Snapshot ELBO.} Consider the snapshot latent $s=(x_t,t)$ and the variational distribution $q^{\mathrm{snap}}(s\mid x_0):=q_t(x_t\mid x_0)$. We define the snapshot predictor $p_0^{\theta,\mathrm{snap}}(x_0\mid s)\deq \mu_\theta(x_t,t)_{x_0}$ from the output of the mean network, and derive the associated snapshot ELBO in \cref{prop:snap_elbo}.

\begin{boxprop}[Snapshot ELBO]\label{prop:snap_elbo}
    Let $x_0\in\cV$, the ELBO is $\log p_0^{\theta,\mathrm{snap}}(x_0) \geq -\cL_{x_0}^{\mathrm{snap}}(\theta)+C^{\mathrm{snap}}_{x_0}$, where
    \[
    \cL_{x_0}^{\mathrm{snap}}(\theta) = \int_0^1\bE_{x_t\sim q_t(\cdot\mid x_0)}\left[-\log \mu_\theta( x_t,t)_{[x_0]}\right]\ud t,
    \]
    and $C^{\mathrm{snap}}_{x_0}$ is independent of $\theta$.
\end{boxprop}

This computable snapshot ELBO boils down to denoising training on $(x_0,x_t,t)$, without requiring the explicit knowledge of $q_t(\cdot\mid x_0)$.It is also well-suited to use with standard time-conditioned bidirectional transformer architectures (e.g., DDiT, \citealt{peebles2023scalable}). Note that \citep{shi2025demystifying} derived the same ELBO expression as a \emph{reweighted} form of the path-wise ELBO (\cref{prop:elbo}), but this equivalence holds only in the masked diffusion setting and for a specific ``simple'' weight.

\begin{algorithm}[htbp]
\caption{GDDS training algorithm}
\label{alg:main_algo}
\begin{algorithmic}[1]
\STATE \textbf{Input:} Training dataset $\bx_0^{(1)},\hdots,\bx_0^{(N_\mathrm{data})}\sim\qdata$
\FOR{$k = 1$ to $N_\mathrm{data}$ (potentially several epochs)}
\STATE Sample $t\sim\mathrm{Unif}[0,1]$
\STATE Sample $\bx_t \sim \smash{\bq_t(\cdot\mid\bx_0^{(k)})}$ with \cref{alg:forward-seqlevel}
\STATE Minimize $-\log(\bmu_\theta(\bx_t,t)_{[\bx_0^{(k)}]})$
\ENDFOR
\STATE \textbf{return} $\bmu_\theta$
\end{algorithmic}
\end{algorithm}

\cref{alg:main_algo} is reminiscent of the general procedure in \citep{bengio2013generalized}, but applied to the corruption process induced by the forward noising diffusion on discrete spaces.

\paragraph{Information-calibration decomposition.}

To make precise the trade-off between \emph{using less information} (a snapshot) and \emph{optimizing better} (lower miscalibration), we compare the expected negative log-likelihood (NLL) of predicting a clean token $x_0\sim q_{\mathrm{data}}$ either from the full forward path $\omega\sim q_{[0,1]}(\cdot\mid x_0)$ or from a randomly sampled snapshot $s=(x_t,t)$ with $t\sim\mathrm{Unif}[0,1]$ independently sampled, through the quantity $\Delta^{NLL}_\theta \coloneqq \bE[-\log p_0^{\theta,\mathrm{snap}}(x_0\mid s)]-\bE[-\log p_0^{\theta,\mathrm{path}}(x_0\mid \omega)]$. The resulting NLL gap admits a clean decomposition into an intrinsic \emph{information path gap} (IPG) and a \emph{calibration gap} (CG), where the calibration is $\Cal_\theta^s\coloneqq\bE[\KL(q(\cdot\mid s)\,\|\,p_\theta(\cdot\mid s))]$.
\begin{boxprop}[Snapshot vs.\ path-wise NLL gap]\label{prop:info_calib_min}
For any conditional predictors $p_0^{\theta,\mathrm{snap}}(\cdot\mid s)$ and $p_0^{\theta,\mathrm{path}}(\cdot\mid \omega)$,
\begin{equation*}
\Delta^{NLL}_\theta=\underbrace{H(x_0\mid s)-H(x_0\mid \omega)}_{\mathrm{IPG}\geq 0}+
\underbrace{\Cal_\theta^s-\Cal_\theta^\omega}_{\mathrm{CG}}.
\end{equation*}
Moreover, $\arg\min_\theta \bE[\cL_{x_0}^{\mathrm{snap}}(\theta)]\!=\!\arg\min_\theta\!\! \Cal_\theta^s$, but $\arg\min_\theta \bE[\cL_{x_0}^{\mathrm{path}}(\theta)]\neq\arg\min_\theta \!\!\Cal_\theta^\omega$ in general.
\end{boxprop}

\begin{figure}    \centering
    \includegraphics[width=\linewidth]{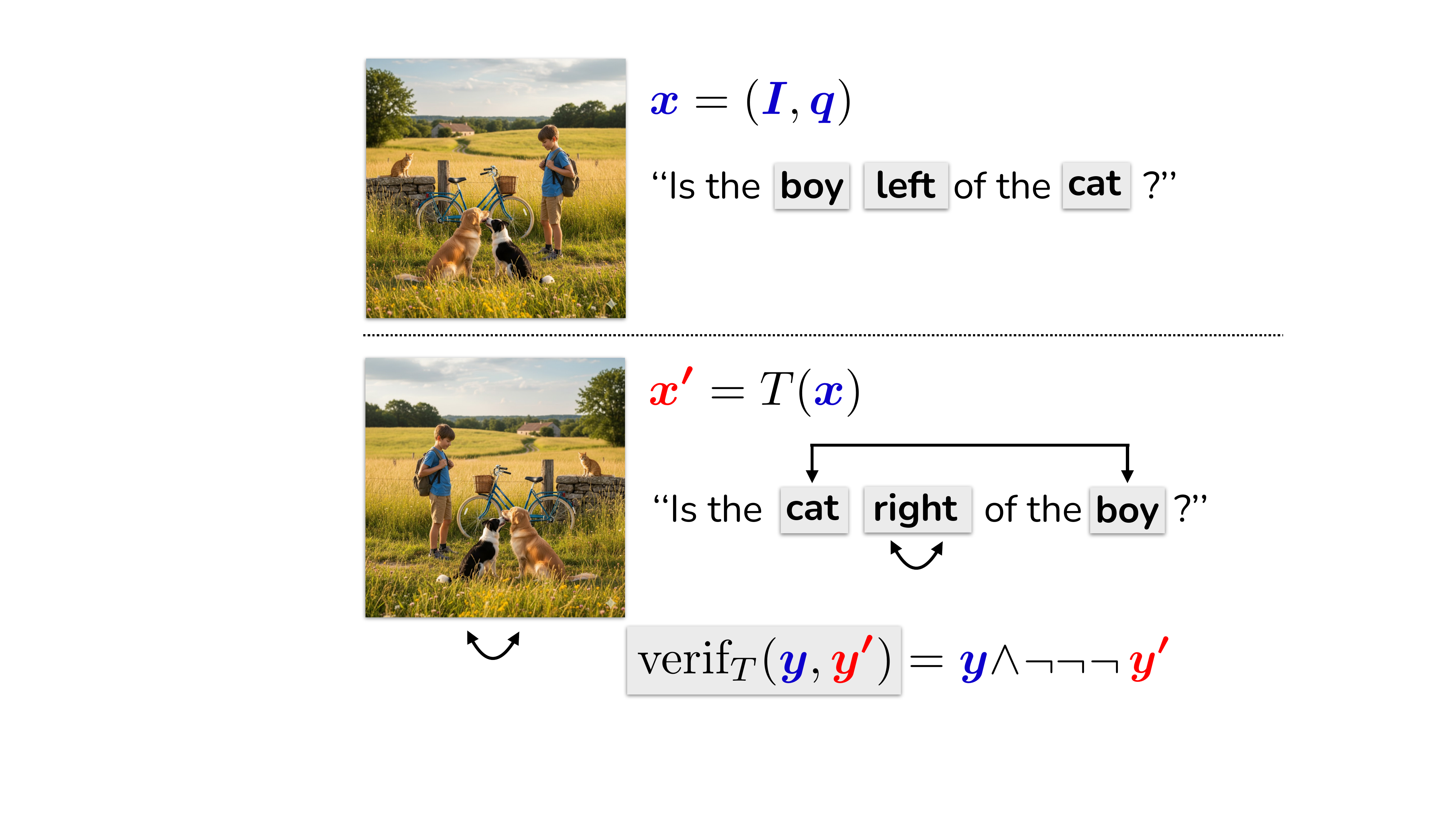}
    \caption{\textbf{Snapshot vs.\ path-wise training.} The forward process corrupts the clean sequence \emph{``My name is David"}. The blue path shows the beginning of the noising trajectory $\textcolor{\PathBlue}{\omega}=\smash{\{(\textcolor{\PathBlue}{x_{t_k}^{(4)}},\textcolor{\PathBlue}{t_k})\}_{k\ge1}}$ of one tracked position ($\ell=4$). Path-wise objectives condition on the entire trajectory $\textcolor{\PathBlue}{\omega}$, whereas our GDDS snapshot objective uses only one random-time observation $\textcolor{\SnapRed}{s}=(\textcolor{\SnapRed}{x_{t^\star}},\textcolor{\SnapRed}{t^\star})$.}     \label{fig:concept_figure}
\end{figure}

This decomposition exposes the core trade-off in replacing path-wise latents $\omega$ by snapshots $s=(x_t,t)$. Discarding the full path induces an intrinsic information loss as $\text{IPG}\geq 0$, which can be compensated by the additional information in $\omega$ when $\text{CG}\leq 0$.
Here, snapshot-latent ELBOs offer a principled trade: they sacrifice some information for an objective that is better aligned with the architecture and easier to optimize, often yielding stronger generative models. In particular, minimizing the snapshot objective enforces $\mu_\theta(x_t,t)_{[x_0]}\approx q(x_0\mid x_t,t)$, which empirically produces better samples as it approximates the correct quantity in \cref{eq:param_old_full}.This distinction between path-wise and snapshot-latent training is illustrated in \cref{fig:concept_figure}.

\section{Experiments}\label{sec:experiments}

We evaluate GDDS on two complementary settings: language modeling and language generation. Additional results and details are deferred to \cref{app:experiments}.

We train character-level models on Text8~\cite{mahoney2024text8} for $1$M steps, and BPE-tokenized models on the widely used OpenWebText (OWT) dataset~\citep{gokaslan2019openwebtext} for $500$k steps. We compare small-model families under matched compute. All retrained models share the same Transformer backbone. Diffusion models use a DDiT backbone with bidirectional attention and time conditioning \citep{peebles2023scalable} ($\approx96$M non-embedding parameters); the autoregressive (AR) baseline uses the same backbone with causal self-attention and no time conditioning ($\approx89$M non-embedding parameters). We retrain AR, a decoder-only Transformer trained with next-token cross-entropy, prior discrete diffusion baselines (UDLM \& MDM) using their respective objectives \citep{schiff2024simple,sahoo2024simple,shi2024simplified,ou2024your}, and our $\GDDS$ snapshot framework with different forward noising processes: $\GDDS$ Absorb (masked), $\GDDS$ Uniform (uniform), and $\GDDS$ Gauss (semantic-informed; \cref{app:forward_gauss}). All reported $\GDDS$ Gauss results use the KNN implementation with $k=64$ neighbors per token.

\begin{table}[ht]
    \centering
    \small
    \setlength{\tabcolsep}{3pt}
    \caption{Bits Per Character ($\downarrow$) on Text8. Baseline results reported from \citep{shi2024simplified}. All models are trained for $1$M steps. Best results per model family are in \textbf{bold}, while best results among retrained models are \underline{underlined}.}
    \begin{tabular}{lr}
    Method &  BPC ($\downarrow$) \\ \midrule
    \textit{Continuous Diffusion} & \\
    Plaid \citep{gulrajani2023likelihood} & $\leq$ 1.48 \\
    BFN \citep{graves2023bayesian} & $\leq$ 1.41\\
    \midrule
    \textit{Any-order Autoregressive} \\
    ARDM \citep{hoogeboom2021autoregressive} & $\leq$ 1.43 \\
    MAC \citep{shih2022training} & $\leq$ 1.40\\
    \midrule
    \textit{Autoregressive} & \\
    IAF/SCF~\citep{ziegler2019latent}  & 1.88\\
    AR Argmax Flow~\citep{hoogeboom2021argmax} & 1.39\\
    Discrete Flow~\citep{tran2019discrete} & \textbf{1.23}\\
    AR~\citep{austin2021structured} & \textbf{1.23}\\
    \midshade AR (retrain)  & \midshade \underline{1.35}\\
    \midrule
    \textit{Uniform Discrete Diffusion} & \\
    Mult. Diffusion \citep{hoogeboom2021argmax} & $\leq$ 1.72\\
    D3PM Uniform \citep{austin2021structured} & $\leq$ 1.61\\
    SEDD Uniform \citep{lou2023discrete} & $\leq$ 1.47 \\
    UDLM \citep{schiff2024simple}  & $\leq$ \textbf{1.44} \\
    \midshade UDLM (retrain)  & \midshade $\leq$ 1.67 \\ 
            \shade $\GDDS$ Uniform (Ours) & \shade $\leq$ \underline{1.50}\\
    \midrule
    \textit{Masked Discrete Diffusion} & \\
    D3PM Absorb \citep{austin2021structured} & $\leq$ 1.45\\
    SEDD Absorb \citep{lou2023discrete} & $\leq$ 1.39 \\
    GenMD4 \citep{shi2024simplified} & $\leq$ 1.34 \\
    MD4 \citep{shi2024simplified} & $\leq$ 1.37 \\
    \midshade MDM (retrain)  & \midshade $\leq$ 1.58 \\
        \shade $\GDDS$ Absorb (Ours) & \shade $\leq$ \underline{\textbf{1.16}}\\
    \bottomrule
    \end{tabular}
\label{tbl:text8}
\end{table}

\paragraph{Semantic-Informed Kernel (SIK).} Following \cref{sec:forward_noise}, one can implement a variety of semantic-informed kernels that depend on the distance between tokens in the embedding space to noise according to words semantic similarities. Let $e_{[x]}$ denote the embedding vector associated with token $x$. Here, we use a Gaussian SIK and, for $x\neq y$, set
\[
F_t^{\mathrm{Gauss}}(x,y)\;\deq\;\frac{\exp\!\left(-\|e_{[x]}-e_{[y]}\|_2^2/\tau(t)\right)}
{\sum_{z\neq y}\exp\left(-\|e_{[z]}-e_{[y]}\|_2^2/\tau(t)\right)},
\]
and $F_t^{\mathrm{Gauss}}(y,y)=0$, where $\tau(t)$ is chosen to increase with $t$ so that the kernel progressively flattens and the forward process approaches the uniform distribution as its limiting distribution. The associated noising algorithm can then be implemented efficiently through \cref{alg:forward-seqlevel} together with either a KNN or a \textsc{KeOps} implementation. In our experiments, $\GDDS$ Gauss is trained with the KNN instantiation using $k=64$ neighbors per token; further implementation details and benchmarks are given in \cref{app:forward_gauss}. More general noising processes can be defined analogously and implemented efficiently thanks to the GDDS framework.

\subsection{Language modeling}

We compute three metrics: Text8 BPC (bits per character) for models trained on Text8 in \cref{tbl:text8}, OWT in-domain validation perplexity on the OWT validation split for OWT-trained models in \cref{tbl:owt_ppl}, and OWT-trained zero-shot perplexity on validation sets of downstream tasks for OWT-trained models in \cref{tbl:owt_zero_shot}. For autoregressive models, these metrics are computed from exact likelihood evaluation. For diffusion models, exact likelihoods are generally not available in closed form; whenever an ELBO is available, we report the corresponding variational upper bound on BPC/perplexity (denoted by $\leq$), using the ELBO associated with the objective the model was trained with.

\begin{table}[ht]
    \centering
    \small
    \setlength{\tabcolsep}{3pt}
    \caption{\textbf{OWT validation perplexity.} Validation perplexity ($\downarrow$) on OWT. Best results are in \textbf{bold}. $^\star$Trained on the WebText dataset. $^\dagger$Result taken from \citep{zhou2025next}. $^\ddagger$Result taken from \citep{sahoo2025diffusion}.}     \begin{tabular}{lcr}
    Method & Training token & PPL ($\downarrow$) \\ \midrule
    \textit{Autoregressive} & \\
    GPT-2$^{\dagger\star}$~\citep{radford2019language} & unknown & 23.40\\
    AR$^\dagger$ & 262B & 16.11\\
    \midshade AR (retrain)  & \midshade 262B & \midshade 20.49\\
    \midrule
    \textit{Uniform Discrete Diffusion} & \\
    SEDD Uniform$^\ddagger$ \citep{lou2023discrete} & 524B & $\leq$ 29.70 \\
    Duo \citep{sahoo2025diffusion} & 524B & $\leq$ 25.20 \\
    \midshade UDLM (retrain)  & \midshade 262B & \midshade $\leq$ 36.82 \\ 
            \shade $\GDDS$ Uniform (Ours) & \shade 262B & \shade $\leq$ \textbf{10.97}\\
    \midrule
    \textit{Masked Discrete Diffusion} & \\
    SEDD Absorb$^\ddagger$ \citep{lou2023discrete} & 524B & $\leq$ 24.10 \\
    GenMD4 \citep{shi2024simplified} & 524B & $\leq$ 21.80 \\
    MDLM \citep{sahoo2024simple} & 327B & $\leq$ 23.00 \\
    \midshade MDM (retrain)  & \midshade 262B & \midshade $\leq$ 31.03 \\
        \shade $\GDDS$ Absorb (Ours) & \shade 262B & \shade $\leq$ \phantom{0}\textbf{8.98}\\
    \midrule
    \textit{General Discrete Diffusion} & \\
        HDLM \citep{zhou2025next} & 131B & $\leq$ 23.25\\
    \shade $\GDDS$ Gauss (Ours) & \shade 262B & \shade $\leq$ \phantom{0}\textbf{7.65}\\
    \bottomrule
    \end{tabular}
\label{tbl:owt_ppl}
\end{table}

\begin{figure}[t]
    \centering
    \includegraphics[width=\linewidth]{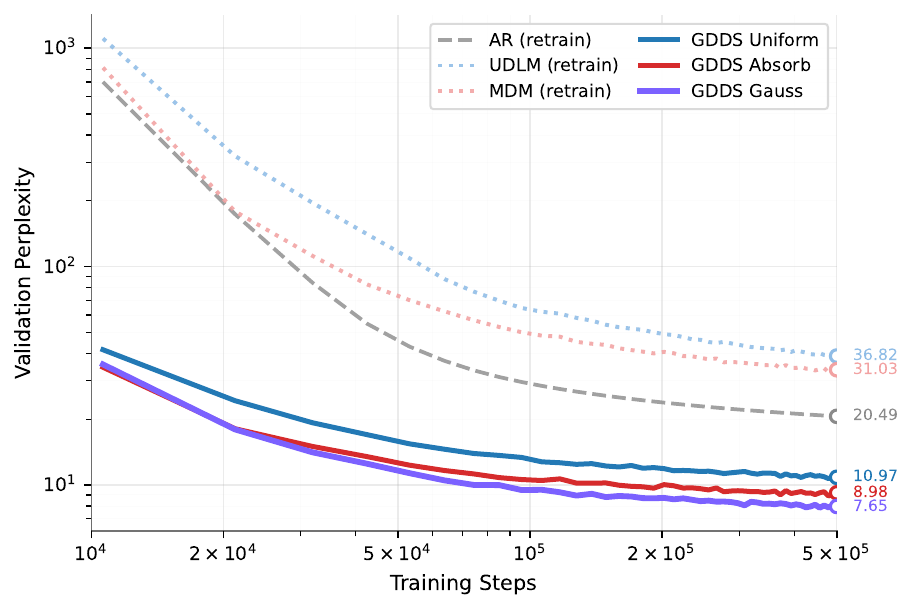}
    \caption{\textbf{OWT training curves.} Evolution of OWT validation perplexity during training for the retrained models reported in \cref{tbl:owt_ppl}. This complements the final numbers in \cref{tbl:owt_ppl} by showing the full optimization trajectory; both axes are shown on logarithmic scales.}
    \label{fig:owt_val_ppl_curve}
\end{figure}

Across both Text8 and OWT, $\GDDS$ substantially improves over prior discrete diffusion baselines under matched compute. On Text8, $\GDDS$ Absorb outperforms the AR baseline for the first time. On OWT, $\GDDS$ yields dramatically tighter variational bounds than retrained UDLM/MDM, and can outperform the matched AR baseline (see \cref{tbl:owt_ppl,fig:owt_val_ppl_curve}). Empirically, we often observe a negative expected NLL gap $\Delta^{\mathrm{NLL}}_\theta<0$, meaning that snapshot objectives improve calibration enough (i.e., $\mathrm{CG}<-\mathrm{IPG}$ in \cref{prop:info_calib_min}). Moreover, because $\mathrm{PPL}=\exp(\mathrm{NLL})$ (see \cref{sec:gen_metrics}), this reduction in NLL can induce a much more pronounced multiplicative drop in perplexity. We find that the choice of forward process matters: on OWT, semantic-informed noise (e.g., $\GDDS$ Gauss) tends to outperform Uniform/Mask by proposing semantically proximal corruptions that are easier to denoise and better aligned with language structure, consistent with prior evidence that semantic similarity improves discrete diffusion modeling \citep{zhou2025next}.

\begin{table}[hbtp]
\caption{\textbf{Zero-shot transfer perplexity.} Zero-shot perplexity ($\downarrow$) of OWT-trained models. Models are evaluated on validation splits of $7$ downstream datasets without additional fine-tuning. Best result per dataset is in \textbf{bold}; second-best is \underline{underlined}.}
\label{tbl:owt_zero_shot}
\centering
\setlength{\tabcolsep}{3pt}
\resizebox{\linewidth}{!}{\begin{tabular}{lccccccc}
\toprule
& PTB & Wikitext103 & LM1B & Lambada  & AG News & Pubmed & Arxiv\\
\midrule
\textit{Autoregressive}& \\
 \midshade AR (retrain) & \midshade 147.90 & \midshade 42.91 & \midshade \underline{81.29} & \midshade 75.93 & \midshade 105.34 & \midshade 79.93 & \midshade 76.29\\
\midrule
\multicolumn{8}{l}{\textit{Masked Discrete Diffusion} (upper bounds $\leq$)} \\
 \midshade MDM (retrain) & \midshade 181.36 & \midshade 45.42 & \midshade 92.58 & \midshade \underline{58.89} & \midshade 123.51 & \midshade 66.46 & \midshade 56.82\\
\shade $\GDDS$ Absorb  & \shade \underline{103.04} &\shade 46.49 & \shade 92.51 &  \shade 60.78 &\shade \underline{101.41} & \shade \underline{62.62} & \shade\underline{53.27}\\
\midrule
\multicolumn{8}{l}{\textit{Uniform Discrete Diffusion} (upper bounds $\leq$)} \\
 \midshade UDLM (retrain) & \midshade 177.26 & \midshade 64.65 & \midshade 112.49 & \midshade 70.38 & \midshade 153.89 & \midshade 70.78 & \midshade 60.35 \\
 \shade $\GDDS$ Uniform  & \shade 115.12 &\shade \underline{42.63} & \shade 108.83 &  \shade 68.92 &\shade 136.24 & \shade 71.69 & \shade 58.83\\
\midrule
\multicolumn{8}{l}{\textit{General Discrete Diffusion} (upper bounds $\leq$)} \\
 \shade $\GDDS$ Gauss  & \shade \textbf{53.65} &\shade  \textbf{34.56} & \shade \textbf{46.06} &  \shade \textbf{38.74} &\shade \textbf{45.94} & \shade \textbf{31.78} & \shade \textbf{28.49}\\
\bottomrule
\end{tabular}
}
\end{table}

Overall, $\GDDS$ improves transfer compared to prior diffusion baselines under matched compute, as shown in \cref{fig:intro_barplot,tbl:owt_zero_shot}. In the masked setting, $\GDDS$ Absorb lowers zero-shot perplexity relative to the retrained MDM on most datasets, indicating better out-of-distribution generalization of the learned denoiser. In the uniform setting, $\GDDS$ Uniform yields large gains over the retrained UDLM across all datasets but one. Most notably, $\GDDS$ Gauss consistently outperforms all baselines by an important margin across every OOD dataset, suggesting that diffusion processes built from semantically structured corruptions may provide a clear generalization advantage.

\subsection{Language generation}

\begin{figure}[ht]
    \centering
    \includegraphics[width=\linewidth]{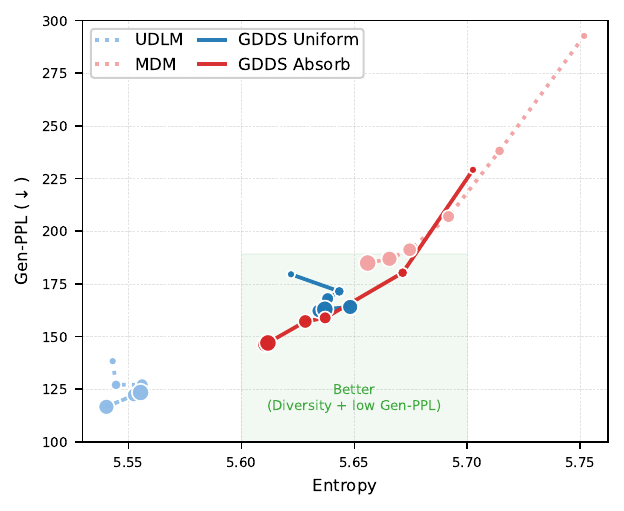}
    \caption{\textbf{Generation quality-diversity tradeoff.} Gen-PPL ($\downarrow$) vs Entropy tradeoff. For $K\in\{32,64,128,256,512,1024\}$ decoding steps, we plot the generative perplexity of $N_{\text{gen}}=256$ unconditional samples under a fixed evaluator (GPT2-large) against their sequence entropy (higher is better). Bubble radius increases with $K$. For reference, the AR baseline achieves Gen-PPL $56.82$ at entropy $5.60$.}
    \label{fig:pareto_gen_entropy}
\end{figure}

We now turn to evaluation generative performance of our models. For numerical stability, we follow \citep{zheng2024masked} and cast logits to \texttt{float64} during sampling. We evaluate $N_{\text{gen}}=256$ unconditional samples from OWT-trained models. In \cref{fig:pareto_gen_entropy}, we consider the Gen-PPL/entropy tradeoff, and report the generative perplexity of unconditional samples under a fixed evaluator (GPT2-large; \citealt{radford2019language}) against their sequence entropy, for multiple decoding budgets $K$. As a reference point for this entropy scale, \citet{zheng2024masked} report that natural OWT text typically falls in the range $5.60$--$5.70$.

Two consistent patterns emerge. First, UDLM attains low Gen-PPL but remains stuck at noticeably lower entropy, suggesting conservative generations with limited diversity. Second, MDM increases entropy as $K$ grows, but this comes with a steep degradation in Gen-PPL at larger budgets, indicating that pushing diversity by running more steps can quickly harm sample quality. In contrast, $\GDDS$ improves the Pareto tradeoff under matched compute. $\GDDS$ Uniform shifts the uniform-diffusion frontier toward higher entropy while keeping Gen-PPL competitive, mitigating the low-diversity behavior of UDLM. More strikingly, $\GDDS$ Absorb yields a strictly better quality/diversity compromise than MDM in the highlighted regime: for comparable entropy, it achieves lower Gen-PPL, and reaches favorable points with substantially fewer decoding steps (smaller bubbles). In particular, at $K=64$ decoding steps, $\GDDS$ Absorb attains a lower Gen-PPL at essentially the same entropy as MDM even when the latter is run with up to $K=1024$ decoding steps, highlighting a large gain in sampling efficiency.

\begin{table}[t]
\centering
\caption{\textbf{Lexical diversity.} Distinct-1/2/3 ($\uparrow$) computed on $N_{\text{gen}}=256$ unconditional samples from OWT-trained models, measuring the fraction of unique $n$-grams among generated texts.}
\small
\setlength{\tabcolsep}{6pt}
\begin{tabular}{lccc}
\textbf{Model} & Dist-1 ($\uparrow$) & Dist-2 ($\uparrow$) & Dist-3 ($\uparrow$) \\
\midrule
\midshade AR (retrain) & 0.10 & 0.57 & 0.88 \\
\midrule
\midshade MDM (retrain) & 0.10 & 0.61 & 0.90	 \\
\shade $\GDDS$ Absorb & 0.10 & 0.60 & 0.89	 \\
\midrule
\midshade UDLM (retrain) & 0.08 & 0.53 & 0.85 \\
\shade $\GDDS$ Uniform & 0.10 & 0.58 & 0.88 \\
\bottomrule
\end{tabular}
\label{tab:diversity}
\end{table}

Next in \cref{tab:diversity}, we consider judge-free quality metrics that do not rely on GPT2-large scoring, and report a diversity statistics (Distinct-$n$). This metric measures the fraction of unique $n$-grams in generated samples (higher is more diverse). Consistent with the low-entropy cluster in \cref{fig:pareto_gen_entropy}, UDLM exhibits the lowest lexical diversity (Distinct-1/2/3). $\GDDS$ Uniform increasing Distinct-1/2/3 over UDLM and matching the strong diversity of masked methods. Overall, Distinct-$n$ corroborates the Pareto analysis: $\GDDS$ increases diversity without incurring the large Gen-PPL penalties observed for MDM at high decoding budgets.

\section{Conclusion}
We introduced Generalized Discrete Diffusion from Snapshots ($\GDDS$), a framework for discrete diffusion models that enables efficient noising processes with arbitrary rate matrix. Our training algorithm relies on a simple loss function based on snapshot samples instead of the entire noising path, and is compatible with standard architectures. GDDS beats previous discrete diffusion models as well as autoregressive models for the first time at this scale on language modeling tasks. Future work may build upon this approach and propose different Semantic-Informed Kernels (SIK) to enforce meaningful noising processes based on similarities between words.

\section*{Impact Statement}
This paper presents work that aims to advance discrete diffusion-based generative modeling in machine learning. There are many potential societal consequences of our work, none of which must be specifically highlighted here.

\ificmlshowauthors
\section*{Acknowledgements}
The authors would like to thank Jules Samaran for precious advice. This project was provided with computing (HPC) and storage resources by GENCI at IDRIS thanks to the grant 2025-AD011017039 on the supercomputer Jean Zay’s V100/A100/H100 partition.
\fi

\bibliography{refs}
\bibliographystyle{icml2026}

\newpage
\appendix
\onecolumn
\textbf{\LARGE Appendix}

\input{appendix/appendix}

\end{document}

%% file: imports/macros.tex
\newcommand{\bE}{\mathbb{E}}
\newcommand{\bN}{\mathbb{N}}
\newcommand{\bP}{\mathbb{P}}
\newcommand{\bR}{\mathbb{R}}

\newcommand{\bj}{\mathbf{j}}
\newcommand{\bx}{\mathbf{x}}

\newcommand{\bp}{\boldsymbol{p}}
\newcommand{\bq}{\boldsymbol{q}}

\newcommand{\bmu}{\boldsymbol{\mu}}
\newcommand{\bomega}{\boldsymbol{\omega}}
\newcommand{\btau}{\boldsymbol{\tau}}
\newcommand{\bcL}{\boldsymbol{\cL}}
\newcommand{\bC}{\mathbf{C}}

\newcommand{\bQ}{\mathbf{Q}}

\newcommand{\bfR}{\mathbf{R}}

\newcommand{\bfJ}{\mathbf{J}}

\newcommand{\cJ}{\mathcal{J}}

\newcommand{\cL}{\mathcal{L}}

\newcommand{\cO}{\mathcal{O}}

\newcommand{\cT}{\mathcal{T}}
\newcommand{\cU}{\mathcal{U}}
\newcommand{\cV}{\mathcal{V}}

\newcommand{\cX}{\mathcal{X}}

\newcommand{\R}{\mathbb{R}}

\renewcommand{\Pr}{\mathbb{P}}

\newcommand{\ud}{\,\mathrm{d}}
\newcommand{\Cal}{\,\mathrm{Cal}}

\newcommand{\sj}{\mathsf{j}}

\newcommand{\unit}{\mathbf{1}}
\newcommand{\deq}{:=}
\newcommand{\KL}{\mathop{\mathrm{KL}}\nolimits}
\newcommand{\CE}{\mathop{\mathrm{CE}}\nolimits}

\newcommand{\softmax}{\mathrm{softmax}}

\newcommand{\dimensionality}{m}

\newcommand{\simplex}{\Delta_{\dimensionality}}
\newcommand{\qdata}{\bq_{\mathrm{data}}}

\newcommand{\horizon}{T}
\newcommand{\Cat}{\mathrm{Cat}}
\newcommand{\Poiss}{\mathrm{Poisson}}

\newcommand{\diag}{\mathrm{diag}}

\newcommand{\GDDS}{\textsc{GDDS}}

\newcommand{\midshade}{\cellcolor{gray!20}}
\newcommand{\shade}{\cellcolor{lavenderblue!30}}



%% file: imports/packages.tex
\usepackage{comment}
\usepackage{nicematrix}
\usepackage{empheq}
\usepackage{multirow}
\usepackage{subfigure}
\usepackage{algorithm}
\usepackage{algorithmic}
\usepackage{bm}
\usepackage{wrapfig}
\usepackage{afterpage}
\usepackage{adjustbox}
\usepackage{amssymb}
\usepackage{enumitem}
\usepackage[table]{xcolor}

\definecolor{rightblue}{RGB}{76,114,176} 
\definecolor{rightorange}{RGB}{221,132,82} 
\definecolor{aliceblue}{rgb}{0.94, 0.97, 1.0} 
\definecolor{darkcerulean}{rgb}{0.03, 0.27, 0.49} 
\definecolor{iris}{rgb}{0.35, 0.31, 0.81} 
\definecolor{carmine}{rgb}{0.59, 0.0, 0.09} 
\definecolor{green(munsell)}{rgb}{0.0, 0.66, 0.47} 
\definecolor{celadon}{rgb}{0.67, 0.88, 0.69} 
\definecolor{bluerow}{rgb}{0.06, 0.3, 0.57} 
\definecolor{lightorange}{RGB}{255, 219, 187} 
\definecolor{lavenderblue}{rgb}{0.8, 0.8, 1.0}
\definecolor{blue(pigment)}{rgb}{0.2, 0.2, 0.6}
\definecolor{blue-violet}{rgb}{0.54, 0.17, 0.89}

\definecolor{bluesepo}{RGB}{30, 60, 110}

\definecolor{redfig}{RGB}{218, 59, 70}
\definecolor{greenfig}{RGB}{58, 133, 75}
\definecolor{bluefig}{RGB}{63, 127, 147}
\definecolor{dpgray}{RGB}{240,240,240} 
\def\dpcolor{dpgray}

\definecolor{bluegray}{rgb}{0.05,0,0.8}
\def\PathBlue{bluegray}
\def\SnapRed{red}

\usepackage{hyperref}
\hypersetup{
  final=true,
  plainpages=false,
  pdfstartview=FitV,
  pdftoolbar=true,
  pdfmenubar=true,
  bookmarksopen=true,
  bookmarksnumbered=true,
  breaklinks=true,
  linktocpage=true,
  colorlinks=true,
  linkcolor=blue-violet, 
  urlcolor=iris, 
  citecolor=darkcerulean, 
  anchorcolor=black
}

\usepackage{amsthm}
\usepackage[capitalize]{cleveref}

\def\propcolor{lavenderblue!30}
\usepackage{mdframed}
\newmdtheoremenv[topline=false, bottomline=false, leftline=false, rightline=false, backgroundcolor=\propcolor,%
innertopmargin=\topskip, splittopskip=\topskip, skipbelow=\baselineskip, skipabove=\baselineskip]{boxthm}{Theorem}[section]

\newmdtheoremenv[topline=false, bottomline=false, leftline=false, rightline=false, backgroundcolor=\propcolor,%
innertopmargin=\topskip, splittopskip=\topskip, skipbelow=\baselineskip, skipabove=\baselineskip]{boxprop}[boxthm]{Proposition}

\newmdtheoremenv[topline=false, bottomline=false, leftline=false, rightline=false, backgroundcolor=\propcolor,%
innertopmargin=\topskip, splittopskip=\topskip, skipbelow=\baselineskip, skipabove=\baselineskip]{boxexample}[boxthm]{Example}
\newmdtheoremenv[topline=false, bottomline=false, leftline=false, rightline=false, backgroundcolor=\propcolor,%
innertopmargin=\topskip, splittopskip=\topskip, skipbelow=\baselineskip, skipabove=\baselineskip]{boxcor}[boxthm]{Corollary}
\newmdtheoremenv[topline=false, bottomline=false, leftline=false, rightline=false, backgroundcolor=\propcolor,%
innertopmargin=\topskip, splittopskip=\topskip, skipbelow=\baselineskip, skipabove=\baselineskip]{boxlem}[boxthm]{Lemma}
\newmdtheoremenv[topline=false, bottomline=false, leftline=false, rightline=false, backgroundcolor=\propcolor,%
innertopmargin=\topskip, splittopskip=\topskip, skipbelow=\baselineskip, skipabove=\baselineskip]{boxdef}[boxthm]{Definition}
\newmdtheoremenv[
  topline=false, bottomline=false, leftline=false, rightline=false,
  backgroundcolor=\dpcolor,
  innertopmargin=\topskip, splittopskip=\topskip,
  skipbelow=\baselineskip, skipabove=\baselineskip
]{designprinciple}[boxthm]{Design Principle}

\BeforeBeginEnvironment{boxprop}{\crefalias{boxthm}{boxprop}}
\BeforeBeginEnvironment{boxlem}{\crefalias{boxthm}{boxlem}}
\BeforeBeginEnvironment{boxcor}{\crefalias{boxthm}{boxcor}}
\BeforeBeginEnvironment{boxdef}{\crefalias{boxthm}{boxdef}}
\BeforeBeginEnvironment{boxexample}{\crefalias{boxthm}{boxexample}}
\BeforeBeginEnvironment{designprinciple}{\crefalias{boxthm}{designprinciple}}

\crefname{boxthm}{Theorem}{Theorems}
\Crefname{boxthm}{Theorem}{Theorems}

\crefname{boxprop}{Proposition}{Propositions}
\Crefname{boxprop}{Proposition}{Propositions}

\crefname{boxexample}{Example}{Examples}
\Crefname{boxexample}{Example}{Examples}

\crefname{boxcor}{Corollary}{Corollaries}
\Crefname{boxcor}{Corollary}{Corollaries}

\crefname{boxlem}{Lemma}{Lemmas}
\Crefname{boxlem}{Lemma}{Lemmas}

\crefname{boxdef}{Definition}{Definitions}
\Crefname{boxdef}{Definition}{Definitions}

\crefname{designprinciple}{Design Principle}{Design Principles}
\Crefname{designprinciple}{Design Principle}{Design Principles}

\newtheoremstyle{mystyle}
  {10pt}
  {10pt}
  {\itshape}
  {}
  {\bfseries}
  {.}
  {.5em}
  {}
\theoremstyle{mystyle}
\newtheorem{thm}{Theorem}[section]

\newtheorem{rmk}[thm]{Remark}


%% file: appendix/appendix.tex
Throughout the appendix, non-bold symbols refer to token-level objects on $\cV$, while bold symbols refer to their sequence-level counterparts on $\cV^n$ for $n\geq 1$.

\section{Time-dependent matrix exponential} \label{app:exp}

\subsection{Transition operator for time-inhomogeneous CTMCs}

Let $t\geq0$, and consider two real $d\times d$ matrices $Q$, $Q_t$ to be time-independent and time-dependent. The matrix exponential of $Q$ is defined by the power series: $\exp(Q)=\sum_{k=0}^\infty Q^k/k!$. When $Q_t$ varies with time, one naturally extends this notion by considering the matrix
$K_{t,s}$ with $0\leq s\leq t$ that solves the following matrix-valued linear ODE:
\[
\frac{\ud}{\ud t}K_{t,s}=Q_tK_{t,s},
\qquad K_{s,s}=I_m.
\]
Its solution can be 
written using the \emph{time-ordered exponential} (or Peano–Baker series) \citep[Thm~1.3.1]{brockett2015finite},
\begin{equation} \label{eq:time_exp}
K_{t,s}=\cT\exp\left(\int_s^t Q_\tau\ud\tau\right),
\end{equation}
where $\cT$ is a time-ordering operator that orders products by decreasing time (later times to the left) as
\[
\cT\exp\left(\int_s^t Q_\tau\ud\tau\right)
= I_d
  + \int_s^t Q_{\tau_1}\ud\tau_1
  + \int_s^t\int_s^{\tau_1} Q_{\tau_1}Q_{\tau_2}\ud\tau_2\ud\tau_1  + \int_s^t\int_s^{\tau_1}\int_s^{\tau_2}
   Q_{\tau_1}Q_{\tau_2}Q_{\tau_3}\ud\tau_3\ud\tau_2\ud\tau_1
  + \cdots,
\]
i.e., the $k$-fold term integrates over $s\leq \tau_k\leq\cdots\le \tau_1\leq t$ the ordered product $Q_{\tau_1}\cdots Q_{\tau_k}$. Note that if the family $\{Q_\tau\}_{\tau\in[s,t]}$ commutes pairwise ($Q_\tau Q_{\tau'}=Q_{\tau'} Q_{\tau}$ for all $\tau,\tau'\in[s,t]$), time ordering is unnecessary and $K_{t,s} = \exp(\int_s^t Q_\tau\ud\tau)$. The time-ordered exponential defined in \cref{eq:time_exp} is useful to describe the fundamental solution to the Kolmogorov forward equation (see \cref{eq:fokker_planck}), which describes the time evolution of a probability vector $q_t$ governed by a (possibly time-dependent) rate matrix $Q_t$. More precisely, the matrix $K_{t,s}$ acts as the linear operator that maps an initial distribution
$q_s$ to its future value $q_t$ as
\[
q_t = K_{t,s}q_s
      = \cT\exp\left(\int_s^t Q_\tau\ud\tau\right)q_s.
\]
For any initial condition $x\in\cX$, the evolution of this point mass under the forward dynamics is obtained by considering $q_0 = \delta_{x}\in\R^m$ at $s=0$, i.e., $q_t(\cdot\mid x) = K_{t,0}\delta_{x}$. Hence, the $x$-th column of $K_{t,s}$ corresponds exactly to the conditional distribution of the CTMC transition matrix at time $t$ given that it was at state $x$ at time $s$, that is,
\[
K_{t,s}(y,x)
= \Pr(x_t = y \mid x_s = x).
\]
In this sense, the time-ordered exponential provides the explicit operator representation of the transition kernel solving the Kolmogorov forward equation.

\subsection{Coordinate-wise rate matrix and product-form marginals}
\label{app:product_form}

Recall that states are sequences $\bx = x^1\hdots x^n \in \cV^n$, and that jumps only modify one token at a time with token-level generator $Q_t^\ell\in\R^{m\times m}$ at position $\ell\in\{1,\hdots,n\}$. Then, the joint rate matrix decomposes as a sum of coordinate-wise generators:
\[
\bQ_t = \sum_{\ell=1}^n \left(I_m \otimes \cdots \otimes Q_t^\ell \otimes \cdots \otimes I_m\right) \in \R^{m^n\times m^n},
\]
where $Q_t^\ell = Q_t$ for all $1\leq \ell\leq n$ acts on coordinate $\ell$ and identities on the others \citep{pauline2025foundations}. Here, $A \otimes B$ denotes the Kronecker product of matrices $A$ and $B$. This structure implies that each coordinate $(x_t^\ell)_{1\leq\ell\leq n}$ evolves as an independent time-inhomogeneous CTMC with rate matrix $Q_t$. In fact, let $q_t(\cdot\mid x^\ell)$ solve the Kolmogorov forward equation:
\[
\frac{\ud}{\ud t} q_t(\cdot\mid x^\ell) = Q_t  q_t(\cdot\mid x^\ell),
\qquad q_0(\cdot\mid x^\ell)=\delta_{x^\ell}.
\]
Define $\tilde{\bq}_t(\cdot\mid\bx) \deq \prod_{\ell=1}^n q_t(\cdot\mid x^\ell)$.
By the product rule and the generator decomposition above, $\tilde{\bq}_t$ solves the joint forward equation
\[
\frac{\ud}{\ud t} \tilde{\bq}_t(\cdot\mid\bx) = \bQ_t \tilde{\bq}_t(\cdot\mid\bx),
\qquad \tilde{\bq}_0(\cdot\mid\bx) = \delta_{\bx}.
\]
Uniqueness of solutions to the forward equation \cref{eq:fokker_planck} yields $\bq_t(\cdot\mid\bx) = \tilde{\bq}_t(\cdot\mid\bx) =\prod_{\ell=1}^n q_t(\cdot\mid x^\ell)$, i.e., the conditional marginal factorizes across coordinates.

\section{Technical results}
\subsection{CTMC rate matrices}\label{app:rate_matrix}
\label{sec:proof_rate_matrix}
\begin{proof}[Proof of \cref{prop:rate_matrix}]
    
Let $t\in(0,1)$ such that $K_t$ is invertible. For each $x\in\cV$, we denote by $v^{(t,x)}$ the solution to linear system $K_t^\top v^{(t,x)}=\dot{K}_t^\top\delta_x$, which exists and is unique. We define a matrix $Q_t\in\R^{m\times m}$ by prescribing its $x$-th column as
\[
Q_t^\top\delta_x\deq v^{(t,x)},\quad \text{for all }x\in\cV.
\]
By construction, for every $x\in\cV$, we have $\dot{K}_t^\top\delta_x= K_t^\top(Q_t^\top\delta_x),$ hence $\dot{K}_t^\top = K_t^\top Q_t^\top$, which is equivalent to $Q_tK_t=\dot{K}_t$ hence $Q_t=\dot{K}_tK_t^{-1}$ since $K_t$ is invertible.

For any $x\neq y\in \cV$, we have
\[
Q_t(y,x) = (Q_t^\top\delta_x)_{[y]} = v^{(t,x)}_{[y]} \geq 0.
\]
Therefore, the off–diagonal entries of $Q_t$ are nonnegative. Since $\Pi_t$ is column–stochastic and $0\leq \alpha_t\leq 1$, each $K_t$ is column–stochastic: $\unit^\top K_t=\unit^\top (\alpha_t I_m + (1 - \alpha_t)\Pi_t)=\unit^\top$. Differentiating with respect to $t$ gives $\unit^\top \dot{K}_t=0$. Using the identity
$\dot{K}_t = K_t Q_t$, we have $0 = \unit^\top \dot{K}_t = \unit^\top K_t Q_t = \unit^\top Q_t$, because $\unit^\top K_t=\unit^\top$. Hence every column of $Q_t$ sums to $0$.
Consequently, for each $x$,
\[
Q_t(x,x) = -\sum_{y\neq x} Q_t(y,x) \leq 0.
\]
Let $p_{x}(t)\deq K_t\delta_x$ be the law at time $t$ when starting from state $x$ at time $0$. Then,
\[
\frac{\ud}{\ud t} p_{x}(t) = \dot{K}_t \delta_x =  Q_tK_t \delta_x = Q_t p_{x}(t),
\]
with $p_{x}(0)=K_0\delta_x=\delta_x$ because $\alpha_0=1$ implies $K_0=I_m$. Thus, $Q_t$ drives the Kolmogorov forward equation, and since its off–diagonals are nonnegative and its columns sum to zero, $Q_t$ is a valid CTMC rate matrix. Combining all previous steps yields the claimed expression $Q_t=\dot{K}_t K_t^{-1}$.
\end{proof}

\begin{boxcor}[GIDD case $\Pi_t = \pi_t\unit^\top$, \citealt{von2025generalized}] \label{cor:gidd}
Let $t\geq0, \pi_t\in\simplex$ and select $\Pi_t = \pi_t\unit^\top\in\R^{d\times d}$, then
\[
Q_t = \frac{\dot{\alpha}_t}{\alpha_t} I_m - \frac{\dot{\alpha}_t}{\alpha_t}\pi_t \unit^\top + (1-\alpha_t)\dot{\pi}_t \unit^\top.
\]
\end{boxcor}

\begin{proof}
We begin by verifying that our regularity assumption on $K_t$ holds. Following the choice $\Pi_t=\pi_t\unit^\top$, we have for all $t\in[0,1]$,
\[
K_t=\alpha_t I_m+(1-\alpha_t)\pi_t\unit^\top,\quad \text{and}\quad \dot{K}_t=\dot{\alpha}_t I_m-\dot{\alpha}_t \pi_t\unit^\top+(1-\alpha_t)\dot{\pi_t}\unit^\top.
\]
Since $K_t$ is a rank-$1$ update to an invertible matrix, the Sherman–Morrison formula~\cite{bartlett1951inverse} ensures that it is invertible and yields the following closed form for its inverse:
\begin{equation} \label{eq:inv_k}
    K_t^{-1}=\frac{1}{\alpha_t}\left(I_m - \frac{(1-\alpha_t)\pi_t \unit^\top}{\alpha_t(1+\frac{1-\alpha_t}{\alpha_t})}\right)=\frac{1}{\alpha_t}(I_m-(1-\alpha_t)\pi_t\unit^\top),\quad t\in(0,1).
\end{equation}
Moreover for all $x\in\cV$ and for all $t\in(0,1)$, one has to verify that $ v^{(t,x)}=(K_t^\top)^{-1}\dot{K}_t^\top\delta_x \geq 0$, i.e. that $(1-\alpha_t)\dot{\pi}_t(x) - \frac{\dot{\alpha_t}}{\alpha_t}\pi_t(x) \geq0$, which is true since $\dot{\alpha_t}\leq 0$ because $\alpha_t$ is a decreasing function. Then, combining \cref{prop:rate_matrix,eq:inv_k} yields
\begin{align*}
Q_t
&=\dot K_t K_t^{-1}=\left(\dot{\alpha}_t I_m - \dot{\alpha}_t\pi_t \unit^\top + (1-\alpha_t)\dot{\pi}_t \unit^\top\right)\frac{1}{\alpha_t}\left(I_m - (1-\alpha_t)\pi_t \unit^\top\right)\\
&=\frac{\dot{\alpha}_t}{\alpha_t}\left(I_m - (1-\alpha_t)\pi_t \unit^\top\right) -\frac{\dot{\alpha}_t}{\alpha_t}\left(\pi_t \unit^\top - (1-\alpha_t)\pi_t \unit^\top \pi_t \unit^\top\right)+ \frac{1-\alpha_t}{\alpha_t}\left(\dot{\pi}_t \unit^\top - (1-\alpha_t)\dot{\pi}_t \unit^\top \pi_t \unit^\top\right).
\end{align*}
Using the identity $\unit^\top \pi_t = 1$, the previous equality simplifies to
\[
Q_t = \frac{\dot{\alpha}_t}{\alpha_t} I_m - \frac{\dot{\alpha}_t}{\alpha_t}\pi_t \unit^\top + (1-\alpha_t)\dot{\pi}_t \unit^\top.
\]
\end{proof}

\begin{proof}[Proof of \cref{prop:mixing_matrix}]
Following the definition of $K_t$ in \cref{eq:main_interpolating_matrix}, for any $t\in[0,1]$, $K_t=\alpha_t I_m+(1-\alpha_t)\Pi_t$. Differentiating the equation with respect to $t$ yields:
\begin{equation} \label{eq:diff_K}
\dot{K}_t = \dot{\alpha}_t I_m + (1-\alpha_t)\dot{\Pi}_t - \dot{\alpha}_t \Pi_t = (1-\alpha_t)\dot{\Pi}_t + \dot{\alpha}_t(I_m-\Pi_t).
\end{equation}
Let $\Pi_t$ be the matrix defined as the unique solution to the following linear matrix-valued ODE:
\[
    \Pi_{0} = I_m - \frac{Q_0}{\dot{\alpha}_0},\quad \text{and}\quad (1-\alpha_t)\dot{\Pi}_t = (\alpha_tQ_t - \dot{\alpha}_t I_m)(I_m - \Pi_t) + Q_t\Pi_t, \quad t\in[0,1],
\] 
which is valid since $0<1-\alpha_t\leq1$ for all $t\in(0,1]$. Inserting the ODE into \cref{eq:diff_K} yields
\begin{align*}
\dot{K}_t
&= (1-\alpha_t)\dot{\Pi}_t + \dot{\alpha}_t(I_m-\Pi_t) = (\alpha_tQ_t - \dot{\alpha}_t I_m)(I_m - \Pi_t) + Q_t\Pi_t + \dot{\alpha}_t(I_m-\Pi_t)\\
&= \alpha_t Q_t - \alpha_t Q_t\Pi_t - \dot{\alpha}_t I_m + \dot{\alpha}_t\Pi_t + Q_t\Pi_t + \dot{\alpha}_t I_m - \dot{\alpha}_t\Pi_t \\
&= \alpha_t Q_t + (1-\alpha_t)Q_t\Pi_t = Q_t(\alpha_t I_m + (1-\alpha_t)\Pi_t) = Q_t K_t.
\end{align*}
For the initial condition, using the Taylor expansion $\alpha_t = 1 + \dot{\alpha}_0 t + o(t)$ as $t\downarrow 0$, we obtain
\[
\lim_{t\downarrow 0} K_t = \lim_{t\downarrow 0} (\alpha_t I_m+(1-\alpha_t)\Pi_t) = I_m.
\]
Hence, for $t\in[0,1]$, $K_t$ satisfies the linear matrix-valued ODE:
\[
\dot{K}_t = Q_t K_t,\quad K_0=I_m,
\]
whose unique solution is $K_t = \cT\exp(\int_0^t Q_s\ud s)$ for all $t\in[0,1]$ \citep[Theorem~1.3.1]{brockett2015finite}. Taking the column associated to the token $x$ on both sides yields $K_t(\cdot,x)=\cT\exp(\int_0^t Q_s\ud s)(\cdot,x)$, so the induced marginals coincide with those of the CTMC. 
\end{proof}

\begin{rmk}
Note that we never explicitly use the expression $\Pi_{0} = I_m - \frac{Q_0}{\dot{\alpha}_0}$ in the proof. We only require $\Pi_{0}$ to be finite; the above choice merely ensures that $\Pi_t$ is uniquely defined as a continuous mixing matrix on $[0,1]$. In fact, we set $\Pi_{0}=I_m-\frac{Q_0}{\dot{\alpha}_0}$ because $1-\alpha_t\to0$ makes the coefficient of $\dot\Pi_t$ in the ODE vanish at $t=0$, and keeping $\dot\Pi_t$ finite requires $(Q_0-\dot{\alpha}_0 I_m)+\dot{\alpha}_0\Pi_{0}=0$.
\end{rmk}

\subsection{Uniformization and exact sampling} \label{app:uniform_sampling}

\begin{proof}[Proof of \cref{prop:mixing_matrix_uniformization}]

Let $(Z_t)_{t\ge0}$ be a CTMC with rate matrix $Q_t = f(t)(F_t-I_m)$, started at $Z_0 = x$. Let $N_t$ be its jump count process. Here, $N_t$ is a non-homogeneous Poisson point process with parameter $\bar{f}(t) = \int_0^tf(s)\ud s$ with marginal distribution $\bP(N_t = k) = \exp(-\bar{f}(t))\bar{f}(t)^k/k!$ for all $k\in\bN$. Conditionally on the event $(N_t=k)$, let $0<T_1<\hdots<T_k<t$ denote the jump times. For each $1\leq r\leq k$, the transition probabilities at a jump are given by the columns of $F_{T_r}$, namely $\bP(Z_{T_r}= i \mid Z_{T_{r-1}}= j, T_r ) = F_{T_r}(i,j)$. Therefore, for any $i,j \in \{1,\hdots,m\}$,\[ 
\bP(Z_{t}= i, N_t = k \mid Z_{0}= x) = \bE\left[(F_{T_k}\hdots F_{T_1})(i,j)\unit_{N_t=k}\mid Z_0 = j\right].
\]
Summing over $k\geq0$ yields
\[
K_t(i,j) = \bP(Z_{t}= i \mid Z_{0}= x) = \sum_{k\geq 0}\bE\left[(F_{T_k}\hdots F_{T_1})(i,j)\unit_{N_t=k}\right],
\]
with the convention that $F_{T_0}\hdots F_{T_1}$ is the identity matrix when $k=0$. Splitting the terms at $k=0$ yields\[
K_t(i,j) = \bP(N_t=0)\unit_{i=j} + \sum_{k\geq 1}\bE\left[(F_{T_k}\hdots F_{T_1})(i,j)\unit_{N_t=k}\right],
\]
Since $\bP(N_t=0) = \alpha_t$, we identify $K_t(i,j) = \alpha_t\unit_{i=j} + (1-\alpha_t) \Pi_t(i,j)$,
where
\[
\Pi_t(i,j) \deq \frac{1}{1-\alpha_t}\sum_{k\geq 1}\bE\left[(F_{T_k}\hdots F_{T_1})(i,j)\unit_{N_t=k}\right] = \bE\left[(F_{T_{N_t}}\hdots F_{T_1})(i,j) \mid N_t\geq 1\right],
\]
as $\bP(N_t\geq 1)= 1-\exp(-\bar{f}(t))=1-\alpha_t$. Note that $\Pi_t$ is a correctly defined mixing matrix. Indeed, for a fixed $k$ and any time sequence $0<t_1<\hdots<t_n<t$, each $F_{T_r}$ is column stochastic and the product $F_{T_k}\hdots F_{T_1}$ as well. Hence, the random matrix $F_{T_{N_t}}\hdots F_{T_1}$ conditioned on $(N_t\geq 1)$ is column stochastic almost surely. Taking the conditional expectation (as a convex combination) preserves this property. Comparing this expression with the interpolating form $K_t=\alpha_t I_m + (1-\alpha_t)\Pi_t$ yields the unique
\[
\Pi_t = \bE\left[(F_{T_{N_t}}\hdots F_{T_1}) \mid N_t\geq 1\right],
\]
as claimed, where uniqueness follows from \cref{prop:mixing_matrix}.
\end{proof}

\begin{rmk}[Exact sampling of $x_t\sim q_t(\cdot\mid x_0)$]
For a given continuous time $t\in[0,1]$, the sampling of $x_t\sim q_t(\cdot\mid x_0)$ can be performed exactly using only a Poisson random variable sampling, uniform random variable sampling, and columns of $F_t$ as follows.
\vspace{-\baselineskip}
\begin{enumerate}[leftmargin=*, itemsep=0pt]
  \item Sample $N_t \sim \Poiss(\bar{f}(t))$.
  \item If $N_t=0$, set $x_t \gets x_0$.
  \item If $N_t\ge 1$, set $z_0 \gets x$ then:
  \begin{enumerate}[itemsep=0pt]
      \item sample ordered jump time $0<T_1<\hdots<T_{N_t}\leq t$ according to the Poisson point process with parameter $\bar{f}(t)$. This can be done by sampling i.i.d. $U_r \sim \cU([0,1])$, set $V_r = \bar{f}^{-1}(U_r\bar{f}(t))$ and take the ordered  statistics of $V_1,\hdots,V_N$ as $T_1,\hdots,T_{N_t}$.
      \item for each $1\leq r\leq N_t$, sample the next state $z_r$ using the $z_{r-1}$-th column of $F_t$ as
\[
\bP(z_{r}=i \mid z_{r-1})= F_{T_r}(i,z_{r-1})\quad\text{for }i=1,\hdots,m.
\]
    \item Set $x_t \gets z_{N_t}$.
  \end{enumerate}
\end{enumerate}
This algorithm produces the conditional law $\bP(x_t=x\mid x_0=y,N_t=k,T_1=t_1,\hdots, T_k =t_k)=F_{t_k}\hdots F_{t_1}(x,y)$.
Marginalizing over $(N_t,T_1,\hdots,T_{N_t})$ yields 
\begin{equation*}
\bP(x_t=x\mid x_0=y)= \sum_{k\geq 0} \bE\left[(F_{T_k}\hdots F_{T_1})(x,y)\unit_{N_t=k}\right]= K_t(x,y).
\end{equation*}
Note that this procedure is exact.
\end{rmk}

\begin{algorithm}[htbp]
\caption{Exact general noising, sequence level (parallel token level)}
\label{alg:forward-seqlevel}
\begin{algorithmic}[1]
\STATE \textbf{Input:} clean sequence $\bx_0 = x_0^1\ldots x_0^n$ of length $n\geq 1$, time $t\in[0,1]$, intensity $\bar{f}(t)$, rate matrix $Q_t$ as in \cref{eq:simplified_rate}
\FOR{$\ell = 1$ to $n$ \textbf{in parallel}}
    \STATE Set $z_0^\ell \gets x_0^\ell$
    \STATE Sample number of jumps $N_t^\ell \sim \Poiss(\bar{f}(t))$
    \STATE Sample and sort the jump times as $T_1^\ell < \ldots < T_{N_t^\ell}^\ell$
    \FOR{$k = 1$ to $\smash{N_t^\ell}$}
        \STATE Sample jump $z_k^\ell \sim F_{T_k^\ell}(\cdot, z_{k-1}^\ell)$
    \ENDFOR
    \STATE Set $x_t^\ell \gets z_{N_t^\ell}^\ell$
\ENDFOR
\STATE \textbf{return} noised sequence $\bx_t = x_t^1\ldots x_t^n$ and per-token jumps $\big\{ (T_k^\ell, z_k^\ell)_{k=1}^{N_t^\ell} \big\}_{\ell=1}^n$
\end{algorithmic}
\end{algorithm}

\subsection{Path-wise parametrization and Evidence Lower Bound (ELBO)}

\subsubsection{Reverse process}\label{app:reverse_process}
 
\paragraph{Forward kernels and true reverse conditionals.}
Fix $0 \le u < t \le 1$ and consider a time-inhomogeneous forward Markov process:
\[
x_0 \to x_u \to x_t,
\quad \text{where}\quad
q(x_0,x_u,x_t)
=
q_{\mathrm{data}}(x_0) q_{u}(x_u\mid x_0) q_{t\mid u}(x_t\mid x_u),
\]
and $q_{t\mid u}$ denotes the forward transition kernel from time $u$ to $t$ (we use the shorthand $q_u \deq q_{u\mid 0}$).
The forward marginal from $x_0$ to time $t$ is
\begin{equation}
\label{eq:marginalization}
q_t(x_t\mid x_0)=\sum_{x_u\in\cV} q_{t\mid u}(x_t\mid x_u) q_{u}(x_u\mid x_0).
\end{equation}
By Bayes' rule, the \emph{true reverse conditional given $x_0$} reads
\begin{equation} \label{eq:true_reverse_given_x0}
q(x_u\mid x_t,x_0) = \frac{q_{t\mid u}(x_t\mid x_u, x_0) q_{u}(x_u\mid x_0)}{\sum_{x_u\in\cV} q_{t\mid u}(x_t\mid x_u) q_{u}(x_u\mid x_0)}
=
\frac{q_{t\mid u}(x_t\mid x_u) q_{u}(x_u\mid x_0)}{q_{t}(x_t\mid x_0)},
\end{equation}
where the second equality uses that the forward process is Markov hence $q_{t\mid u}(x_t\mid x_u, x_0)=q_{t\mid u}(x_t\mid x_u)$ and the marginalization \eqref{eq:marginalization}. 
Finally, conditioning on the observed endpoint $x_t$, the \emph{true reverse kernel} is the mixture
\begin{equation} \label{eq:true_reverse_mixture}
q(x_u\mid x_t)
=
\sum_{x_0\in\cV} q(x_u\mid x_t,x_0) q_{0\mid t}(x_0\mid x_t),
\end{equation}
where $q_{0\mid t}(x_0\mid x_t)$ is the exact forward posterior of $x_0$ given the snapshot $s=(x_t,t)$.

\paragraph{Plug-in Bayes and realizability.}
At inference time $x_0$ is unknown while $(x_t,t)$ is observed. Let $\mu_\theta(\cdot\mid x_t,t)$ be a neural predictor (e.g. a transformer head; \citealt{vaswani2017attention}) that outputs a distribution over $x_0\in\cV$ given $(x_t,t)$.
We define the \emph{plug-in reverse transition} by mixing the exact conditional \eqref{eq:true_reverse_given_x0} with $\mu_\theta$ as
\begin{equation} \label{eq:plugin_reverse_kernel}
p_\theta(x_u\mid x_t,t)
\deq
\sum_{x_0\in\cV} q(x_u\mid x_t,x_0) \mu_\theta(x_0\mid x_t,t).
\end{equation}
This parametrization is widely used in the literature \citep{austin2021structured,campbell2022continuous,sahoo2024simple,shi2024simplified,von2025generalized,zhou2025next}. It is interesting, as it verifies the following lemma.

\begin{boxlem}[Realizable limit]\label{lem:plugin_realizable}
If $\mu_\theta(\cdot\mid x_t,t)=q_{0\mid t}(\cdot\mid x_t)$, then $p_\theta(\cdot\mid x_t,t)$ equals the true reverse kernel:
\begin{equation*}
p_\theta(x_u\mid x_t,t)=q(x_u\mid x_t).
\end{equation*}
\end{boxlem}

\begin{proof}
If $\mu_\theta(\cdot\mid x_t,t)=q_{0\mid t}(\cdot\mid x_t)$, then substituting into \cref{eq:plugin_reverse_kernel} gives
\[
p_\theta(x_u\mid x_t,t)
=
\sum_{x_0\in\cV} q(x_u\mid x_t,x_0) q_{0\mid t}(x_0\mid x_t)
=
q(x_u\mid x_t),
\]
where the last equality is exactly \eqref{eq:true_reverse_mixture}.
\end{proof}

A first crucial insight is that, $\mu_\theta(\cdot\mid x_t,t)$ should be train to approach the posterior $q_{0\mid t}(\cdot\mid x_t)$, so that $p_\theta(x_u\mid x_t,t)$ approaches $q(x_u\mid x_t)$.

\paragraph{CTMC view and factorization of the reverse dynamics.}
For an infinitesimal step $u=t-\epsilon$ with $\epsilon\downarrow 0$, the true reverse transition admits the standard ``no-jump + jump'' expansion:
\[
q(x_{t-\epsilon}\mid x_t)
=
(1-\epsilon r_{[x_t]}(t))\delta_{x_t}
+\epsilon r_{[x_t]}(t)R_t(\cdot,x_t)
+o(\epsilon),
\]
where $r_{[x_t]}(t)$ is the true reverse exit rate from state $x_t$ and $R_t(\cdot,x_t)\in\Delta_m$ is the true jump destination distribution.
Equivalently, the reverse generator factorizes as
\[
\overline{Q}_t = (R_t-I_m) \diag(r_1(t),\dots,r_m(t)),
\]
i.e.,
\[
\overline{Q}_t(y,x)
=
\begin{cases}
r_{[x]}(t) R_t(y,x), & y\neq x,\\
-r_{[x]}(t), & y=x.
\end{cases}
\]
This factorization separates \emph{when} the chain jumps (through $r_{[x]}(t)$) from \emph{where} it jumps (through $R_t$).

\paragraph{Why the plug-in Bayes kernel entangles ``where'' and ``when''.}
\cref{eq:plugin_reverse_kernel} induces a \emph{time-inhomogeneous} reverse kernel whose
short-time behaviour is governed by a $\theta$-dependent generator.
Indeed, for an infinitesimal step $u=t-\epsilon$ with $\epsilon\downarrow 0$, the induced kernel
$p_\theta(x_{t-\epsilon}\mid x_t,t)$ admits a first-order expansion of the form
\[
p_\theta(x_{t-\epsilon}\mid x_t,t)
=
(1-\epsilon r_{[x_t]}^{\theta}(t))\delta_{x_t}
+\epsilon r_{[x_t]}^{\theta}(t) A_t^\theta(\cdot, x_t)
+o(\epsilon),
\]
where \emph{both} the effective exit rate $r_{[x_t]}^\theta(t)$ and the jump destination distribution
$A_t^\theta(\cdot, x_t)\in\Delta_m$ depend on $\theta$ through $\mu_\theta(\cdot\mid x_t,t)$.
Equivalently, the induced reverse generator $\smash{\overline{Q}_t^\theta}$ factorizes as
\begin{equation}\label{eq:old_param}
\overline{Q}_t^\theta = (A_t^\theta-I_m) \diag(r_1^\theta(t),\dots,r_m^\theta(t)),
\end{equation}
so that learning $\mu_\theta$ implicitly learns \emph{both} a jump kernel and a time-dependent clock.
In other words, the Bayes plug-in construction couples \emph{where} the chain jumps (through $A_t^\theta(\cdot,x_t)$)
and \emph{when} it jumps (through $r_{[x_t]}^\theta(t)$), which complicates the path-wise ELBO and its optimization:
the event-level objective contains gradients through both the destination cross-entropy term and the rate/normalization term,
rather than isolating a single $\theta$-dependent jump component.

\paragraph{Jump-states parametrization: learn only the jump kernel, fix the exit rates.}
To mirror the true CTMC factorization, we instead parameterize only the jump destinations while prescribing the exit rates by a diffusion schedule.
Concretely, we choose a nonnegative schedule $r_{[x]}(t)\ge 0$ independently of $\theta$ and define the neural reverse generator
\begin{equation}\label{eq:new_param}
\overline{Q}^{\theta}_t=(R_t^\theta-I_m) \diag(r_1(t),\dots,r_m(t)),
\quad \text{where} \quad
R_t^\theta(y,x)=\sj_\theta(x,t)_{[y]},
\end{equation}
such that
\[
\overline{Q}^{\theta}_t(y,x)
=
\begin{cases}
r_{[x]}(t) \sj_\theta(x,t)_{[y]}, & y\neq x,\\
-r_{[x]}(t), & y=x.
\end{cases}
\]
Equivalently, for $u=t-\epsilon$ the associated reverse kernel satisfies the first-order expansion:
\begin{equation*}
p_\theta(x_{t-\epsilon}\mid x_t)
=
(1-\epsilon r_{[x_t]}(t))\delta_{x_t}
+\epsilon r_{[x_t]}(t) \sj_\theta(x_t,t)_{x_{t-\epsilon}}
+o(\epsilon).
\end{equation*}
Compared to the plug-in Bayes transition in \cref{eq:plugin_reverse_kernel}, the network now controls only \emph{where} the chain jumps (via $\sj_\theta(x_t,t)$), while \emph{when} it jumps is entirely prescribed by the schedule $r_{[x_t]}(t)$.

\subsubsection{Proof of \cref{prop:elbo}}\label{app:elbo}

The following proposition is valid for any general forward CTMC and its parametrized time reversal. It has been established and rewritten in several previous works (e.g., \citealt{campbell2022continuous,shi2024simplified,von2025generalized,zhou2025next}), and we restate it with our notations.

\begin{boxprop}[General Path-wise ELBO, \citealt{campbell2022continuous}]\label{prop:previous_elbo}
    Let $x_0\in\cV$, the ELBO is given by $\log p_0^{\theta,\mathrm{path}}(x_0) \geq -\tilde{\cL}_{x_0}^{\mathrm{path}}(\theta)+\tilde{C}^{\mathrm{path}}_{x_0}$, where $\tilde{C}^{\mathrm{path}}_{x_0}$ is independent of $\theta$ and
    \[
    \tilde{\cL}_{x_0}^{\mathrm{path}}(\theta) =\int_0^1 \bE_{x_t\sim q_t(\cdot\mid x_0)}\left[-\overline{Q}_t^\theta(x_t,x_t) +\sum_{y\neq x_t}Q_{t}(y,x_t)(-\log R_t^\theta(x_t,y))\right]\ud t.
    \]
\end{boxprop}

Let us apply \cref{prop:previous_elbo} to our jump-states parametrization detailed in \cref{sec:path_wise}. In our case,  $\overline{Q}_t^\theta(x_t,x_t) = - r_{[x_t]}(t)$ is $\theta$-independent (see \cref{eq:jump_parametrization}), we can discard it from the $\theta$-dependent part, and define 
\begin{equation} \label{eq:l_path}
    \cL_{x_0}^{\mathrm{path}}(\theta) \deq\int_0^1 \bE_{x_t\sim q_t(\cdot\mid x_0)}\left[\sum_{y\neq x_t}Q_{t}(y,x_t)(-\log R_t^\theta(x_t,y))\right]\ud t,
\end{equation} 
and $C^{\mathrm{path}}_{x_0} \deq \tilde{C}^{\mathrm{path}}_{x_0} - \int_0^1 \bE_{x_t\sim q_t(\cdot\mid x_0)}[r_{[x_t]}(t)]\ud t$. We can then rewrite the ELBO as
$\log p_0^{\theta,\mathrm{path}}(x_0) \geq -\cL_{x_0}^{\mathrm{path}}(\theta)+C^{\mathrm{path}}_{x_0}$.

We now define the following quantities similarly to \citep[Lem.~2]{shi2024simplified}.

\begin{boxdef}[True conditional reverse jump kernel $R_t^{x_0}$ and conditional reverse rate $r^{x_0}_x(t)$]\label{app:def:Rx0}
Fix $x_0\in\cV$ and let $q_t(\cdot\mid x_0)$ be the forward marginal at time $t$.
For $x\in\cV$ with $q_t(x\mid x_0)>0$, define the \emph{conditional reverse jump kernel} $R_t^{x_0}(\cdot,x)$ by, for all $y\neq x$,
\[
R_t^{x_0}(y,x)\deq\frac{q_t(y\mid x_0)}{q_t(x\mid x_0)} \frac{Q_t(x,y)}{r_{[x]}^{x_0}(t)},
\qquad\text{where }
r_{[x]}^{x_0}(t)\deq\sum_{y\neq x}\frac{q_t(y\mid x_0)}{q_t(x\mid x_0)} Q_t(x,y).
\]
If $q_t(x\mid x_0)=0$, the value of $R_t^{x_0}(\cdot,x)$ is irrelevant under $x_t\sim q_t(\cdot\mid x_0)$ and is set arbitrarily to $0$.
\end{boxdef}

To conclude the proof, we provide the following lemma which gives two clean rewriting of the $\theta$-dependent part $\cL_{x_0}^{\mathrm{path}}(\theta)$.

\begin{boxlem}\label{lem:expected_ce}
The quantity $\cL_{x_0}^{\mathrm{path}}(\theta)$ defined in \cref{eq:l_path} satisfies the following identities:
\begin{align*}
    \cL_{x_0}^{\mathrm{path}}(\theta)
    &= \int_0^1 \bE_{x_t\sim q_t(\cdot\mid x_0)}\left[\sum_{y\neq x_t}\frac{q_t(y\mid x_0)}{q_t(x_t\mid x_0)}Q_{t}(x_t,y)(-\log R_t^\theta(y,x_t))\right]\ud t\\
    &= \int_0^1\bE_{x_t\sim q_t(\cdot\mid x_0)}\big[r_{[x_t]}^{x_0}(t)\CE(R_t^{x_0}(\cdot, x_t),R_t^\theta(\cdot, x_t))\big]\ud t.
\end{align*}
Here, $\CE(R_t^{x_0},R_t^\theta)\big|_{x_t}$ denotes the cross-entropy between $R_t^{x_0}(\cdot,x_t)$ and $R_t^\theta(\cdot,x_t)$. $R_t^{x_0}$ and $r^{x_0}(t)$ denote respectively the conditional reverse jump kernel and its associated exit rate introduced in \cref{app:def:Rx0}.
\end{boxlem}

\begin{proof}
Let $t\in[0,1]$, we start from the definition of $\cL_{x_0}^{\mathrm{path}}(\theta)$ in \cref{eq:l_path} and expands the the expectation term inside the time integral to obtain
\begin{align*}
\bE_{x_t\sim q_t(\cdot\mid x_0)}\Big[\sum_{y\neq x_t}Q_t(y,x_t)\big(-\log R_t^\theta(x_t,y)\big)\Big]
&=\sum_{x\in\cV} q_t(x\mid x_0)\sum_{y\neq x}Q_t(y,x)\big(-\log R_t^\theta(x,y)\big)\\
&=\sum_{x\in\cV}\sum_{y\neq x} q_t(x\mid x_0) Q_t(y,x)\big(-\log R_t^\theta(x,y)\big).
\end{align*}
Swapping the variables names $(x,y)\mapsto(y,x)$ inside the double sum yields
\begin{align}
\bE_{x_t\sim q_t(\cdot\mid x_0)}\Big[\sum_{y\neq x_t}Q_t(y,x_t)\big(-\log R_t^\theta(x_t,y)\big)\Big]
&=\sum_{x\in\cV}\sum_{y\neq x} q_t(y\mid x_0) Q_t(x,y)\big(-\log R_t^\theta(y,x)\big) \label{eq:inner_sum_Q_t}\\
&=\sum_{x\in\cV} q_t(x\mid x_0)\sum_{y\neq x}\frac{q_t(y\mid x_0)}{q_t(x\mid x_0)} Q_t(x,y)\big(-\log R_t^\theta(y,x)\big) \notag\\
&=\bE_{x_t\sim q_t(\cdot\mid x_0)}\Big[\sum_{y\neq x_t}\frac{q_t(y\mid x_0)}{q_t(x_t\mid x_0)}Q_t(x_t,y)\big(-\log R_t^\theta(y,x_t)\big)\Big].\notag
\end{align}
Integrating over $t$ gives the first identity.

For the second identity, we note that for $y\neq x$, \cref{app:def:Rx0} can be restated as
\[
\frac{q_t(y\mid x_0)}{q_t(x\mid x_0)}Q_t(x,y)=r_{[x]}^{x_0}(t) R_t^{x_0}(y,x).
\]
Plugging this into the inner sum in \cref{eq:inner_sum_Q_t} yields, for each $x\in\cV$,
\begin{equation*}
\sum_{y\neq x}\frac{q_t(y\mid x_0)}{q_t(x\mid x_0)}Q_t(x,y)\big(-\log R_t^\theta(y,x)\big)
= r_{[x]}^{x_0}(t)\sum_{y\neq x} R_t^{x_0}(y,x)\big(-\log R_t^\theta(y,x)\big)= r_{[x]}^{x_0}(t) \CE\left(R_t^{x_0}(\cdot,x), R_t^\theta(\cdot,x)\right).
\end{equation*}
Taking the expectation over $x_t\sim q_t(\cdot\mid x_0)$ and integrating over $t\in[0,1]$ provides the desired quantity:
\[
\cL_{x_0}^{\mathrm{path}}(\theta)
=\int_0^1 \bE_{x_t\sim q_t(\cdot\mid x_0)}\Big[r_{[x_t]}^{x_0}(t) \CE\left(R_t^{x_0}(\cdot,x_t),R_t^\theta(\cdot,x_t)\right)\Big]\ud t.
\]
\end{proof}

A notable simplification comes from the jump-state parametrization. As shown in \cref{lem:expected_ce}, the only $\theta$-dependent contribution is a weighted cross-entropy over reverse jump matrices. By contrast, in the SEDD formulation of \citet{lou2023discrete}, the ELBO contains two $\theta$-dependent components: (i) a linear score term and (ii) a cross-entropy term. In the mean parametrization of \citet{campbell1909study} (recalled in \cref{prop:previous_elbo}), the linear score term is replaced by a $\theta$-dependent reverse-rate term. In both settings, training must therefore fit \emph{both} jump destinations and a separate $\theta$-dependent factor that controls the time-change, which adds burden and variance. Our jump-state parametrization removes this extra requirement: it keeps the reverse rate fixed and concentrates learning on the core problem (predicting \emph{where} the chain jumps) via a single cross-entropy objective. The jump-state parametrization is closely related to the approach of \citet{amin2025masking}. However, we do not condition the path-wise generative model on an explicit event schedule; instead, we keep the same joint distribution defining the generative model as in prior work.

\subsubsection{Proof of \cref{prop:campbell}}

Let $x_0\sim q_0$ and define the \textit{forward jump flow measure} on the space $\Omega\deq [0,1]\times\cV\times\cV$ as
\[
\Lambda^{x_0}(\ud t, \ud y, \ud x) \deq q_t(x\mid x_0)Q_t(y,x)\unit_{\{y\neq x\}}\ud t.
\]
We then rewrite the expression of $\cL_{x_0}^{\mathrm{path}}(\theta)$ in \cref{eq:l_path} as
\begin{equation} \label{eq:l_path_2}
\cL_{x_0}^{\mathrm{path}}(\theta) = \int_0^1 \bE_{x\sim q_t(\cdot\mid x_0)}\left[\sum_{y\neq x}Q_{t}(y,x)(-\log R_t^\theta(x,y))\right]\ud t= \int_{\Omega} -\log R_t^\theta(x,y)\Lambda^{x_0}(\ud t, \ud y, \ud x).
\end{equation}
We now introduce $\cJ^{x_0}(\ud t,\ud y,\ud x) \deq \sum_{k=1}^{N_1} \delta_{(T_k,z_{k},z_{k-1})}(\ud t,\ud y,\ud x)$, the jump random measure on $\Omega$ of the forward CTMC started at $x_0$. Here, $N_1$ and $(T_k,z_{k})_{1\leq k \leq N_1}$ are the output of \cref{alg:forward-tokenlevel} with inputs $x_0$, $t=1$ and $Q_t$. Equivalently, this notation means that for any measurable set $A\subset\Omega, \cJ^{x_0}(A) = \sum_{k=1}^{N_1} \unit_{\{(T_k,z_{k},z_{k-1})\in A\}}$. Its intensity measure is exactly $\Lambda^{x_0}$ as for any measurable set $A\subset\Omega$, we have $\bE[\cJ^{x_0}(A)]=\Lambda^{x_0}(A)$. To see this, it suffices to verify the identity on a rectangle $A=(a,b]\times \{y\}\times\{x\}$ with $0\leq a < b\leq 1$ and $x\neq y \in \cV$, and extend to all measurable sets by a monotone class argument. For such a rectangle, we have
\[
\bE[\cJ^{x_0}(A)] = \bE\left[\sum_{k=1}^{N_1} \unit_{\{T_k\in(a,b],z_{k}=y,z_{k-1}=x\}}\right] \eqqcolon \bE\left[C^{x\to y}_{(a,b]}\right],
\]
where $(x_t)_{t\in[0,1]}$ is the forward CTMC started at $x_0$, and $C^{x\to y}_{(a,b]}$ is the number of jumps between states $x\to y$ in the time interval $(a,b]$. Let $r\geq 1$, we consider a uniform partition of $(a,b]$ as $t_0=a<\ldots<b=t_{r}$ with step-size $\Delta = (b-a)/r$. The Markov property and the generator give $\bE[C^{x\to y}_{(t_i,t_{i+1}]}]=\bP(X_{t_i}=x)\left(Q_{t_i}(y,x)\Delta+o(\Delta)\right)$. Summing over $0\leq i\leq r-1$ yields a Riemann sum along with a remainder term of the form $\sum_{i=0}^{r-1} o(\Delta)$, which vanishes as the mesh goes to $0$. Therefore, $\bE[C^{x\to y}_{(a,b]}]=\int_a^b \bP(x_t=x)Q_t(y,x)\ud t$ and, since $\bP(x_t=x) = q_t(x\mid x_0)$, we obtain $\bE[\cJ^{x_0}(A)] = \Lambda^{x_0}(A)$.
Hence, Campbell's formula \citep[Prop.~$2.7$]{last2018lectures} states that for any nonnegative measurable function $g:\Omega\to\R_{\geq 0}$,
\begin{equation} \label{eq:campbell_appl}
\bE_{(T_k,z_{k},z_{k-1})_k}\left[\sum_{k=1}^{N_1} g(T_k,z_{k},z_{k-1})\right] = \int_{\Omega} g(t,y,x)\Lambda^{x_0}(\ud t,\ud y,\ud x).
\end{equation}
To simplify the notations, we introduce $\omega=\{N_1,(z_k,T_k)_{k=1}^{N_1}\}$ and $q_{[0,1]}(\cdot\mid x_0)$ such that $\bE_{(T_k,z_{k},z_{k-1})_k} = \bE_{\omega\sim q_{[0,1]}(\cdot\mid x_0)}$. Choosing $g(t,y,x) = -\log R_t^\theta(x,y)\geq0$ and combining \cref{eq:l_path_2,eq:campbell_appl} yield 
\[
\cL_{x_0}^{\mathrm{path}}(\theta) =  \bE_{\omega\sim q_{[0,1]}(\cdot\mid x_0)}
\left[
\sum_{k=1}^{N_1}-\log R_{T_k}^\theta(z_{k-1},z_k)
\right]=\bE_{\omega\sim q_{[0,1]}(\cdot\mid x_0)}
\left[
\sum_{k=1}^{N_1}-\log j_\theta(z_k,T_k)_{z_{k-1}}
\right],
\]
by definition of $j_\theta$.

For an expression at the sequence level (of size $n\geq 1)$, let $\bx_0 = x_0^1\dots x_0^n\sim\qdata$. Using $\bq_t(\cdot\mid\bx_0) = \prod_{\ell=1}^n q_t(\cdot\mid x^\ell_0)$ implies that that the sequence-level measure is the superposition of the $n$ independent token jump measure $\cJ^{x_0^\ell}$ (i.e., $\bfJ^{\bx_0}=\sum_{\ell=1}^n \cJ^{x_0^\ell}$). The linearity of the expectation implies that
\[
\cL(\theta) = \bE_{\bomega\sim \bq_{[0,1]}(\cdot\mid\bx_0)}
\left[
\sum_{\ell=1}^n\sum_{k=1}^{N^\ell_1}
-\log R_{T_k^\ell}^\theta(z_{k-1}^\ell,z_k^\ell)\right],
\]
where $\bomega=\{\omega^\ell\}_{\ell=1}^n\sim \bq_{[0,1]}(\cdot\mid\bx_0)$ with $\omega^\ell=\{N^\ell_1,(T_k^\ell,z_k^\ell)_k\}$, and $\bq_{[0,1]}(\cdot\mid\bx_0)=\prod_{\ell=1}^n q_{[0,1]}(\cdot\mid x_0^\ell)$.

\subsubsection{Recovering the masked diffusion loss}\label{app:md4_mdlm}

For masked diffusion, \citep{ou2024your,sahoo2024simple,shi2024simplified} concurrently discovered a simplified expression of the \citep{campbell2022continuous}-ELBO, that collapses with \cref{prop:elbo}. This is due to the fact that masked diffusion makes the reverse exit rates independent of $\theta$ \citep{amin2025masking}. Hence, the seminal parametrization of \citep{austin2021structured,campbell2022continuous} collapses with our jump-states parametrization in this special case. To our knowledge, this property is only true for masked diffusion, and other attempts of generalization from the parametrization of \citep{austin2021structured,campbell2022continuous} led to more complicated losses than the weighted cross-entropy given by our jump-states parametrization (see e.g., \citealt{von2025generalized,zhou2025next}).

Consider the forward generator $Q_t^{\mathrm{absorb}}=f(t)(F^{\mathrm{absorb}}-I_m)$ with a single mask state $\texttt{[MASK]}$ (i.e., $F_t = F^{\mathrm{absorb}} = \delta_\texttt{[MASK]}\unit^\top$, or equivalently $\Pi_t = \Pi = \delta_\texttt{[MASK]}\unit^\top$), so that $Q_t^{\mathrm{absorb}}(\texttt{[MASK]}, j)=f(t)$ for $j\neq \texttt{[MASK]}$, and no other off-diagonal is nonzero. Remember that $q_t(x_t\mid x_0)=\alpha_t\delta_{x_t=x_0}+\big(1-\alpha_t\big)\delta_{x_t=\texttt{[MASK]}}$. This means $x_t \sim q_t(\cdot\mid x_0)$ is either $x_t=x_0$ or $x_t=\texttt{[MASK]}$. In this rank-1 absorbing case, the mean parametrization and the jump-states parametrization given by \cref{eq:jump_parametrization} coincide. Hence, $\theta$ only governs the transitions, while the scheduling is fixed by $\alpha_t$. In fact, one has the following lemma.

\begin{boxlem} \label{lem:param_masked_diff}
In masked diffusion, the mean parametrization and the jump-states parametrization introduced in \cref{eq:jump_parametrization} coincide. In particular, the (conditional) reverse rate is independent of $\theta$ and given by
\[
r_{[x]}^{x_0}(t) =r_{[x]}(t)=
\begin{cases}
\dfrac{-\dot{\alpha}_t}{1-\alpha_t}, & x=\texttt{[MASK]},\\
0, & x\neq \texttt{[MASK]}.
\end{cases}
\]
\end{boxlem}

\begin{proof}

Recall that
\begin{equation} \label{eq:transition_mask}
q_t(y\mid x_0)=
\begin{cases}
\alpha_t\delta_{y=x_0}, & y\neq \texttt{[MASK]},\\
1-\alpha_t, & y=\texttt{[MASK]},
\end{cases}
\end{equation}
and that \cref{app:def:Rx0} defined the conditional reverse rate at state $x$ as
\[
r^{x_0}_{[x]}(t)=\sum_{y\neq x} Q_t^{\mathrm{absorb}}(x, y)\frac{q_t(y\mid x_0)}{q_t(x\mid x_0)}.
\]

\paragraph{Computation for $x=\texttt{[MASK]}$.}
Only the forward edges $y\to \texttt{[MASK]}$ ($y\neq \texttt{[MASK]}$) are nonzero, i.e., $Q_t(\texttt{[MASK]}, y)=f(t)$. Therefore,
\[
r^{x_0}_{\texttt{[MASK]}}(t)
=\sum_{y\neq \texttt{[MASK]}} Q_t^{\mathrm{absorb}}(\texttt{[MASK]}, y)\frac{q_t(y\mid x_0)}{q_t(\texttt{[MASK]}\mid x_0)}
=\sum_{y\neq \texttt{[MASK]}} f(t)\frac{\alpha_t\delta_{y=x_0}}{1-\alpha_t}
=f(t)\frac{\alpha_t}{1-\alpha_t} =\frac{-\dot{\alpha}_t}{1-\alpha_t},
\]
as $f(t) = -\dot{\alpha}_t/\alpha_t$ by definition of the mixing rate (cf. \cref{sec:efficient_forward}).

\paragraph{Computation for $x\neq \texttt{[MASK]}$.} There is no forward transition onto non-mask token $x\neq \texttt{[MASK]}$ ($Q_t(x, y)=0$ for all $y\in\cV$), hence $r^{x_0}_{[x]}(t)=0$ for all $i\neq \texttt{[MASK]}$.

\paragraph{From $r^{x_0}_{[x]}(t)$ to $r_{[x]}(t)$.}
Marginalizing over $x_0\sim q_0$ yields
\[
q_t(\texttt{[MASK]})=\sum_{x_0\in\cV} q_0(x_0)\,q_t(\texttt{[MASK]}\mid x_0)=\sum_{x_0\in\cV} q_0(x_0)\,(1-\alpha_t)=1-\alpha_t,
\]
where we used \cref{eq:transition_mask}. Hence, $q_t(\texttt{[MASK]}\mid x_0)=q_t(\texttt{[MASK]})=1-\alpha_t$, which implies that 
\[\sum_{y\neq \texttt{[MASK]}}q_t(y\mid x_0)=\alpha_t=\sum_{y\neq \texttt{[MASK]}}q_t(y).\] 
Considering $r_{[x]}(t)$, we can repeat the proof of $r^{x_0}_{[x]}(t)$ since $r_{[x]}(t)=\sum_{y\neq x} Q_t^{\mathrm{absorb}}(x, y)\frac{q_t(y)}{q_t(x)}$, and obtain 
\[r_{[x]}(t)=
\begin{cases}
\dfrac{-\dot{\alpha}_t}{1-\alpha_t}, & x=\texttt{[MASK]},\\
0, & x\neq \texttt{[MASK]},
\end{cases}=r^{x_0}_{[x]}(t).\]
\end{proof}

We can now prove that our ELBO in \cref{prop:elbo} derived from the jump states parametrization coincides with the ELBO of masked diffusion models derived from the mean parametrization in \citep{ou2024your,sahoo2024simple,shi2024simplified}.

\begin{boxprop}[Recovering the MDM loss]
For the case of masked diffusion, our ELBO in \cref{prop:elbo} coincides with the MDM loss:
\[
\cL_{x_0}^{\mathrm{path}}(\theta) =\int_0^1 \frac{-\dot{\alpha}_t}{1-\alpha_t}\bE_{x_t\sim q_t(\cdot\mid x_0)}\left[\unit_{\{x_t = \texttt{[MASK]}\}}(-\log \mu_\theta( x_t,t))_{x_0}\right]\ud t.
\]
\end{boxprop}

\begin{proof}
\cref{lem:param_masked_diff} directly implies that $A_t^\theta=R_t^\theta$ in \cref{eq:old_param,eq:new_param}, so the mean and jump parameterization collapses in the masked diffusion case ($\mu_\theta=\sj_\theta$). Recovering the MDM loss is pretty straightforward from the expression $\cL_{x_0}^{\mathrm{path}}(\theta) =\int_0^1 \bE_{x_t\sim q_t(\cdot\mid x_0)}\left[\sum_{y\neq x_t}Q_{t}^{\mathrm{absorb}}(y,x_t)(-\log \mu_\theta(y,t)_{x_t})\right]\ud t$. In fact,
\begin{enumerate}[topsep=0pt]
    \item if $x_t = x_0$, then the token is still clean, and we have $\sum_{y\neq x_t}Q_{t}^{\mathrm{absorb}}(y,x_t)(-\log \mu_\theta(y,t)_{x_t}) = Q_t^{\mathrm{absorb}}(\texttt{[MASK]},x_0)(-\log \mu_\theta(\texttt{[MASK]},t)_{x_0}) = f(t)(-\log \mu_\theta(\texttt{[MASK]},t)_{x_0})$;
    \item if $x_t = \texttt{[MASK]}$, the token is masked. In this case, $Q_t^{\mathrm{absorb}}(y,\texttt{[MASK]}) =0$ for all $y\neq \texttt{[MASK]}$. Hence, $\sum_{y\neq x_t}Q_{t}^{\mathrm{absorb}}(y,x_t)(-\log \mu_\theta(y,t)_{x_t})=0$.
\end{enumerate}
Combining these cases and using $\bP(x_t=x_0\mid x_0) =\alpha_t$ and $\bP(x_t=\texttt{[MASK]}\mid x_0) =1-\alpha_t$, we obtain $\cL_{x_0}^{\mathrm{path}}(\theta) =\int_0^1 \alpha_tf(t)(-\log \mu_\theta(\texttt{[MASK]},t)_{x_t})\ud t = \int_0^1 -\dot{\alpha}_t(-\log \mu_\theta(\texttt{[MASK]},t)_{x_t})\ud t$, where we used $\dot{\alpha}_t = -\alpha_tf(t)$. This is the simplest form of the loss, but is usually written differently by leveraging the identity $\bE_{x_t\sim q_t(\cdot\mid x_0)}[\unit_{\{x_t = \texttt{[MASK]}\}}(-\log \mu_\theta(x_t,t)_{x_0})] = (1-\alpha_t)(-\log \mu_\theta(\texttt{[MASK]},t)_{x_0})$. Finally, we recover the MDM loss as
\[
\cL_{x_0}^{\mathrm{path}}(\theta) =\int_0^1 \frac{-\dot{\alpha}_t}{1-\alpha_t}\bE_{x_t\sim q_t(\cdot\mid x_0)}\left[\unit_{\{x_t = \texttt{[MASK]}\}}(-\log \mu_\theta( x_t,t))_{x_0}\right]\ud t.
\]
\end{proof}

\subsection{Snapshot parametrization and Evidence Lower Bound (ELBO)}

\begin{proof}[Proof of \cref{prop:snap_elbo}]
    Consider the snapshot latent $s=(x_t,t)$ and the variational distribution $q^{\mathrm{snap}}(s\mid x_0)=\rho(t)q_t(x_t\mid x_0)$, where $\rho$ is the uniform distribution over $[0,1]$ for simplicity (i.e., $\rho(t) =1$ for all $t\in[0,1]$). The snapshot marignal is then $q^{\mathrm{snap}}(s) = \bE_{x_0\sim q_{\mathrm{data}}}[q^{\mathrm{snap}}(s\mid x_0)]=\bE_{x_0\sim q_{\mathrm{data}}}[q_t(x_t\mid x_0)] = q_t(x_t)$. We define the snapshot predictor $p_0^{\theta,\mathrm{snap}}(x_0\mid s)\coloneqq \mu_\theta(x_t,t)_{x_0}$ from the output of the mean network. This defines a latent-variable model with joint probability
\[
p_0^{\theta,\mathrm{snap}}(x_0,s) = p_0^{\theta,\mathrm{snap}}(x_0\mid s)q^{\mathrm{snap}}(s).
\]
A standard ELBO derivation (see e.g., \citealt[Thm.~2.1.1]{lai2025principles}) consists of applying Jensen's inequality as follows,
\[
\log p_0^{\theta,\mathrm{snap}}(x_0) = \log \bE_{s\sim q^{\mathrm{snap}}(\cdot\mid x_0)}\left[\frac{p_0^{\theta,\mathrm{snap}}(x_0,s)}{q^{\mathrm{snap}}(s\mid x_0)}\right]\geq \bE_{s\sim q^{\mathrm{snap}}(\cdot\mid x_0)} \left[\log \frac{p_0^{\theta,\mathrm{snap}}(x_0,s)}{q^{\mathrm{snap}}(s\mid x_0)}\right].
\]
Expanding the right hand-side using $p_0^{\theta,\mathrm{snap}}(x_0,s) = p_0^{\theta,\mathrm{snap}}(x_0\mid s)q^{\mathrm{snap}}(s)$ yields
\[
\log p_0^{\theta,\mathrm{snap}}(x_0) \geq -\cL_{x_0}^{\mathrm{snap}}(\theta)+C^{\mathrm{snap}}_{x_0},
\]
where $C^{\mathrm{snap}}_{x_0} \coloneqq -\bE_{s\sim q^{\mathrm{snap}}(\cdot\mid x_0)} \left[\log \frac{q^{\mathrm{snap}}(s\mid x_0)}{q^{\mathrm{snap}}(s)}\right]$ is independent of $\theta$ and
\[
\cL_{x_0}^{\mathrm{snap}}(\theta) \coloneqq \bE_{s\sim q^{\mathrm{snap}}(\cdot\mid x_0)} \left[-\log {p_0^{\theta,\mathrm{snap}}(x_0\mid s)}\right]= \int_0^1\bE_{x_t\sim q_t(\cdot\mid x_0)}\left[-\log \mu_\theta( x_t,t)_{x_0}\right]\ud t.
\]
\end{proof}

We now move on to the proof of \cref{prop:info_calib_min}. Consider a clean token $x_0\sim q_{\mathrm{data}}$. The path-wise latent $\omega=\{N_1,(z_k,T_k)_{k=1}^{N_1}\}\sim q_{[0,1]}(\cdot\mid x_0)$ is the full forward CTMC path on $[0,1]$ and $s=(x_t,t)$ is the snapshot latent from this path. This means that we have sampled $t\sim\rho(t)$ (with $\rho$ being the uniform distribution over $[0,1]$) independently of $x_0$, and extracted the associated token $x_t$ from $\omega$. This token is defined as follows : there exists some $k_t\in\{1,\hdots,N_1\}$ such that $t\in(T_{k_t-1},T_{k_t}]$, and $x_{t}=z_{k_t-1}$ (where the edge case $k_t=1$ is covered by setting $T_0=0$ and $z_0=x_0$). More formally, if $\Omega$ is the path space and $S$ the snapshot space, there exists a measurable map $\Psi:[0,1]\times\Omega\to S$ such that $\Psi(t,\omega)=s$. For $v\in \{s,\omega\}$, the joint law of both generative models can be written as $q(x_0,v) = q_{\mathrm{data}}(x_0)q(v\mid x_0)$ and the corresponding posterior as $q(x_0\mid v)=q(x_0,v)/q(v)$. Let $p_\theta^{\mathrm{snap}}(\cdot\mid s)$ and $p_\theta^{\mathrm{path}}(\cdot\mid \omega)$ be any conditional predictors. We define the expected NLL gap as $\Delta^{NLL}_\theta = \bE[-\log p_0^{\theta,\mathrm{snap}}(x_0\mid s)]-\bE[-\log p_0^{\theta,\mathrm{path}}(x_0\mid \omega)]$ and the calibration error as $\Cal_\theta^v=\bE[\KL(q(x_0\mid v)\,\|\,p_\theta(\cdot\mid v))]$. Let us start by introducing a simple lemma.

\begin{boxlem}[Expected NLL decomposition]\label{lem:expected_nll}
Let $(X,V)\sim q(\cdot,\cdot)$ be any pair, and let $p_\theta(\cdot\mid v)$ be any conditional predictor for $v\in \{s,\omega\}$. Then,
\[
\bE_{(X,V)\sim q(\cdot,\cdot)}\left[-\log p_\theta(X\mid V)\right] = H(X\mid V) + \bE_V\left[\KL\left(q(\cdot\mid V)\,\|\,p_\theta(\cdot\mid V)\right)\right].
\]
\end{boxlem}

\begin{proof}
After conditioning on $V=v$, we obtain
\[
\bE\left[-\log p_\theta(X\mid V)\right] = \bE_{V}\left[\bE_{X\sim q(\cdot\mid V)}[-\log p_\theta(X\mid V)]\right].
\]
For each fixed $v\in\cV$, the inner expectation is the cross-entropy between $q(\cdot\mid v)$ and $p_\theta(\cdot\mid v)$:
\[
\bE_{X\sim q(\cdot\mid v)}[-\log p_\theta(X\mid v)]=H\left(q(\cdot\mid v),p_\theta(\cdot\mid v)\right) = H\left(q(\cdot\mid v)\right) + \KL\left(q(\cdot\mid v)\,\|\,p_\theta(\cdot\mid v)\right).
\]
Taking expectation over $V$ gives
\[
\bE[-\log p_\theta(X\mid V)] = \bE_V[H(q(\cdot\mid V))] + \bE_V[\KL(q(\cdot\mid V)\,\|\,p_\theta(\cdot\mid V))].
\]
However, $\bE_V[H(q(\cdot\mid V))]=H(X\mid V)$ by definition of conditional entropy.
\end{proof}

\begin{proof}[Proof of \cref{prop:info_calib_min}]
Applying \cref{lem:expected_nll} twice with $V=s$ and $V=\omega$, and subtracting the two yields the following decomposition:
\[
\Delta^{NLL}_\theta=\underbrace{H(x_0\mid s)-H(x_0\mid \omega)}_{\mathrm{IPG}}+
\underbrace{\Cal_\theta^s-\Cal_\theta^\omega}_{\mathrm{CG}},
\]
where $\Cal_\theta^s\coloneqq \bE[\KL(q(\cdot\mid s)\,\|\,p_\theta(\cdot\mid s))]$. Note that, by measurability of $\Psi:[0,1]\times\Omega\to S$, the data processing inequality states that the mutual information can only decrease, i.e., $I(x_0;\omega)\geq I(x_0;s)$, or equivalently $H(x_0\mid s) \geq H(x_0\mid \omega)$, which implies that $\mathrm{IPG}\geq 0$.

\paragraph{Snapshot minimizer.}
The snapshot loss is given by
\begin{equation} \label{eq:snap_min}
\cL^{\mathrm{snap}}_{x_0}(\theta)=\int_0^1 \bE_{x_t\sim q_t(\cdot\mid x_0)}\left[-\log \mu_\theta(x_t,t)_{x_0}\right]\ud t=\bE\left[-\log p_\theta^{\mathrm{snap}}(X_0\mid s)\,\big|\,X_0=x_0\right].
\end{equation}
Note that it would be weighted by $\rho(t)$ inside the time integral if $\rho$ was chosen differently from the uniform density.
Averaging \cref{eq:snap_min} over $x_0\sim q_{\mathrm{data}}$ yields exactly the unconditional expected snapshot NLL:
\[
\bE_{x_0\sim q_{\mathrm{data}}}[\cL^{\mathrm{snap}}_{x_0}(\theta)]=\bE_{x_0\sim q_{\mathrm{data}}}\bE_{s\sim q^{\mathrm{snap}}(\cdot\mid x_0)}[-\log p_\theta^{\mathrm{snap}}(x_0\mid s)].
\]
Following \cref{lem:expected_nll}, $\bE_{x_0\sim q_{\mathrm{data}}}[\cL^{\mathrm{snap}}_{x_0}(\theta)]=H(x_0\mid s)+\Cal_\theta^s$, where the term $H(x_0\mid s)$ depends only on the data distribution and the forward process, and not on $\theta$. Therefore
\[
\arg\min_\theta \bE[\cL^{\mathrm{snap}}_{x_0}(\theta)]=\arg\min_\theta \Cal_\theta^s.
\]

\paragraph{Path-wise minimizer.} Contrary to a snapshot ELBO, a path-wise ELBO (i.e., a diffusion ELBO; see \citealt[Thm.~2.2.3]{lai2025principles}) contains an additional diffusion term that depends on $\theta$. This makes the model calibrates the local conditionals $z_{k-1}\mid (z_{k},T_k)$ instead of $x_0\mid \omega$. This can be clearly seen here with the Campbell form of the path-wise ELBO. Indeed, recall that $\Cal_\theta^\omega \deq \bE_\omega\left[\KL\left(q(x_0\mid \omega)\,\|\,p_\theta(\cdot\mid \omega)\right)\right]$ measures calibration of a predictor of the \emph{initial token} $x_0$ given the \emph{full path} $\omega$.
In contrast, the Campbell path-wise objective
\[
\cL_{x_0}^{\mathrm{path}}(\theta)
=
\bE_{\omega\sim q_{[0,1]}(\cdot\mid x_0)}
\left[\sum_{k=1}^{N_1}-\log \sj_\theta(z_k,T_k)_{z_{k-1}}\right]
\]
is the negative log-likelihood of predicting the \emph{previous state} $z_{k-1}$ at each jump time from the event $(z_k,T_k)$.
More precisely, using the other form given by \cref{prop:elbo},
\[
\cL_{x_0}^{\mathrm{path}}(\theta)=\int_0^1 \bE_{x_t\sim q_t(\cdot\mid x_0)}\left[
r_{x_t}^{x_0}(t)\,\CE\left(R_t^{x_0}(\cdot,x_t),R_t^\theta(\cdot,x_t)\right)
\right]\ud t,
\]
hence, up to a $\theta$-independent term,
\[
\cL_{x_0}^{\mathrm{path}}(\theta)=\text{const}+\int_0^1 \bE_{x_t\sim q_t(\cdot\mid x_0)}\left[r_{x_t}^{x_0}(t)\,\KL\left(R_t^{x_0}(\cdot,x_t)\,\|\,R_t^\theta(\cdot,x_t)\right)
\right]\ud t.
\]
Therefore, minimizing $\bE[\cL_{X_0}^{\mathrm{path}}(\theta)]$ calibrates the \emph{local reverse jump kernel} $R_t^\theta(\cdot,x)$, not the posterior predictor $p_\theta(x_0\mid \omega)$. Since $\Cal_\theta^\omega$ concerns a different conditional distribution (namely $x_0\mid \omega$), the minimizers do not coincide in general.
\end{proof}

\section{Experimental details}\label{app:experiments}

\subsection{Experimental setting}

We train $\GDDS$ models on Text8 and OWT with bf16 precision (including the loss), global batch size 512, AdamW with learning rate $3.5\times 10^{-4}$, weight decay $10^{-2}$, $(\beta_1,\beta_2)=(0.9,0.95)$, gradient clipping at 1.0, and EMA with decay 0.9999, for 1M optimizer steps, validating every 10k steps. All runs are executed on a single node with 4$\times$ NVIDIA H100 GPUs using DDP. More details are given in \cref{tab:training configuration}.

\begin{table}[hbtp]
\centering
\small
\caption{$\GDDS$ Training configuration}
\begin{tabular}{ll}
\toprule
\textbf{Category} & \textbf{Setting} \\
\midrule
Sequence length & 256 for Text8, 1024 for OWT \\
Hidden size / heads & 768 / 12 \\
MLP ratio / dropout & 4 / 0.1 \\
Time conditioning & AdaLN\\
Layers & 12 \\
Optimizer & AdamW ($\beta_1=0.9,\beta_2=0.95,\epsilon=10^{-8}$), weight decay$=0.01$ \\
Learning rate & $3.5\times 10^{-4}$ \\
LR schedule & Constant warmup for the 2500 first steps \\
Precision & bf16 (training and loss precision) \\
EMA & 0.9999 \\
Batch size & Global batch size 512 (eval global batch size 512) \\
Gradient clipping & 1.0 \\
Max steps & 1,000,000 \\
Eval & validation every 10,000 steps \\
Noise & log-linear \\
Hardware & 1 node, 4$\times$ NVIDIA H100 GPUs; 4 tasks per node (DDP) \\
\bottomrule
\end{tabular}
\label{tab:training configuration}
\end{table}

\paragraph{Text8 dataset.}
We train models on Text8 for 1 million optimizer steps. While results in \cref{tbl:text8} are reported from \citep{shi2024simplified}, we additionally retrained three baselines: a standard autoregressive Transformer (AR), MDM \citep{sahoo2024simple,shi2024simplified,ou2024your} and UDLM \citep{schiff2024simple} using the same training recipe as for $\GDDS$ (optimizer, learning rate, schedule, batch size, precision, number of steps, and evaluation protocol). The only architectural difference is that the AR baseline uses no time conditioning and causal attention. Under this unified setup, we obtain $1.35$, $1.58$ and $1.67$ BPC for AR, MDM and UDLM respectively (see \cref{tbl:text8}). Notably, these values are worse than the corresponding numbers reported in \cref{tbl:text8} in \citep{shi2024simplified}, suggesting that cross-paper comparisons are sensitive to training details (e.g., optimization and implementation choices) and experimental conditions (e.g., hardware and system-level settings). Note that our AR result exactly matches the one found by \citep{hoogeboom2021autoregressive}. Since $\GDDS$ Absorb still outperforms these baselines by a large margin under our unified setup, we do not emphasize this mismatch further but mention it for completeness.

\paragraph{OpenWebText dataset.}
We follow the same unified protocol on OWT, training all models for $500$k optimizer steps at sequence length $1024$ (see \cref{tab:training configuration}), resulting in a total of $262$B training tokens. We tokenize OpenWebText using the GPT-2 tokenizer, then concatenate documents and chunk the resulting stream into sequences of length $1024$, inserting an \texttt{<|endoftext|>} token between consecutive documents. Since OpenWebText has no official validation split, we reserve the last $100$k tokens for validation. Similarly than Text8, we retrain the matched-compute baselines (AR, MDM, and UDLM) with the same optimizer, learning-rate schedule, batch size, precision, EMA, and evaluation pipeline as our $\GDDS$ variants. Under this controlled setup, we observe again that our retrained baselines do not reproduce the perplexities reported in prior papers (see \cref{tbl:owt_ppl}), highlighting the sensitivity of OWT results to implementation details and system-level factors (e.g., data processing, optimization hyperparameters, and hardware/software stacks). For this reason, we primarily rely on our retrained baselines for fair comparisons, while still reporting prior-work numbers when available. Importantly, $\GDDS$ yields substantially tighter likelihood bounds and stronger generation quality under the same protocol, and the main conclusions remain unchanged despite the mismatch with previously reported figures.

In \cref{tbl:owt_zero_shot}, we evaluate zeroshot perplexity by taking the models trained on OpenWebText and evaluating likelihoods on the validation splits of 7 datasets: Penn Tree Bank (PTB; \citealt{marcus1993building}), Wikitext103 \citep{merity2016pointer}, One Billion Word Language Model Benchmark (LM1B; \citealt{chelba2013one}), Lambada \citep{paperno2016lambada}, AG News \citep{zhang2015character}, and scientific papers (Pubmed and Arxiv subsets; \citealt{cohan2018discourse}). Since the zeroshot datasets have different conventions for sequence segmentation, we wrap sequences to 1024 but do not
add \texttt{<|endoftext|>} tokens in between sequences. 

\subsection{Semantic-Informed Kernel (SIK)}\label{app:forward_gauss}

Recall that $m$ is the vocabluary size. Let $E\in\mathbb{R}^{m\times d}$ be a fixed embedding table, where $e_i\in\mathbb{R}^d$ denote the embedding vector of the token number $i\in\{1,\dots,m\}$.
Let $\{\rho_i\}_{i=1}^m$ be positive local bandwidths (e.g., $\rho_i$ is the squared distance from $e_i$ to its $k$-th nearest neighbor). We fix (i) a decaying profile $\eta:[0,\infty)\to\mathbb{R}_+$, (ii) a distance on the embedding space $d_{\mathrm{emb}}:\mathbb{R}^d\times\mathbb{R}^d\to[0,\infty)$, and (iii) a temperature schedule $\tau:[0,1]\to(0,\infty)$ that controls the amount of mixing (small $\tau(t)$ yields a near-identity kernel; large $\tau(t)$ yields a flatter kernel). Following self-tuning Diffusion Maps~\citep{zelnik2004self}, we define the (symmetric) affinity as
\begin{equation}
\label{eq:Kemb_def}
W_t(x,y)\deq\eta\!\left(\frac{d_{\mathrm{emb}}(e_{[x]},e_{[y]})^2}{\tau(t)\sqrt{\rho_{[x]}\rho_{[y]}}}\right),
\quad \text{for }x\neq y\in\cV, \quad W_t(x,x)=0.
\end{equation}
Hence, it is possible to define the Semantic-Informed Kernel by normalizing the affinity:
\[
F_t^{\mathrm{SIK}}(x,y)
\deq
\frac{W_t(x,y)}{\sum_{z\neq y}W_t(z,y)},
\qquad x\neq y,
\qquad F_t^{\mathrm{SIK}}(y,y)=0.
\]
For the Gaussian metric, $d_{\mathrm{emb}}(e_{[x]},e_{[y]})=\|e_{[x]}-e_{[y]}\|_2^2$, while for the cosine metric, $d_{\mathrm{emb}}(e_{[x]},e_{[y]})=1-\langle \bar e_{[x]}, \bar e_{[y]}\rangle$ after normalizing embeddings to unit norm. The corresponding CTMC rate matrix is $Q_t^{\mathrm{SIK}} \deq f(t)\left(F_t^{\mathrm{SIK}}-I_m\right)$, with $f$ such that $\alpha_t = \exp(-\bar{f}(t))$ as in \cref{prop:mixing_matrix_uniformization}. The embeddings are extracted from GPT-2 \citep{radford2019language}, where $d=768$ and $m =50257$.

\paragraph{Gaussian and cosine SIKs.}
In the experiments, we consider two concrete choices for the affinity defining $F_t^{\mathrm{SIK}}$. The first is the Gaussian SIK, based on squared Euclidean distance in embedding space. The second is a cosine SIK, obtained by first normalizing embeddings to unit norm and then using the cosine distance $d_{\mathrm{emb}}(e_{[x]},e_{[y]}) = 1-\langle \bar e_{[x]}, \bar e_{[y]}\rangle$. The Gaussian version favors tokens that are close in Euclidean geometry, while the cosine version instead depends only on the angle between embeddings. The main text notation $F_t^{\mathrm{Gauss}}$ corresponds precisely to this Gaussian choice of the SIK affinity.

\paragraph{Two implementations: KNN and \textsc{KeOps}.}
There are then two ways to implement $F_t^{\mathrm{SIK}}$ in practice. The first is a KNN implementation, where for each source token $y$ we precompute a sparse neighbor set $\mathcal N_k(y)$ containing only its top-$k$ nearest candidate neighbors in embedding space (in our benchmark, $k=64$). In that case, we use the sparse approximation
\[
\begin{aligned}
F_t^{\mathrm{KNN}}(x,y)
&=
\left(1-\lambda(t)\right)
\frac{\exp\!\left(-\frac{d_{\mathrm{emb}}(e_{[x]},e_{[y]})}{\varepsilon\,\sqrt{\rho_{[x]}\rho_{[y]}}}\right)\unit_{x\in\mathcal N_k(y)}}
{\sum_{z\in\mathcal N_k(y)}\exp\!\left(-\frac{d_{\mathrm{emb}}(e_{[z]},e_{[y]})}{\varepsilon\,\sqrt{\rho_{[z]}\rho_{[y]}}}\right)}
\;+\;
\lambda(t)\frac{\unit_{x\neq y}}{m-1}
\end{aligned}
\]
so sampling takes place only on this sparse candidate set, which makes the method very fast in practice. The second is a \textsc{KeOps} implementation, denoted $F_t^{\mathrm{KeOps}}$, where we do not build a sparse graph and do not materialize the dense $m\times m$ kernel either; instead, we evaluate the entries of $F_t^{\mathrm{SIK}}$ on-the-fly with blockwise GPU reductions. Thus, KNN should be viewed as a sparse approximation of $F_t^{\mathrm{SIK}}$, whereas \textsc{KeOps} is the dense lazy implementation of the same normalized kernel.

\paragraph{Why this is \textsc{KeOps}-friendly.}
For the dense implementation $F_t^{\mathrm{KeOps}}$, we only specify the symbolic pairwise map
\[
(\ell,j) \longmapsto\ \eta\!\left(\frac{d_{\mathrm{emb}}(e_\ell,e_j)^2}{\tau(t)\sqrt{\rho_\ell\rho_j}}\right).
\]
Thus, we let \textsc{KeOps} generate fused GPU kernels that compute the needed sums \emph{on-the-fly}. This avoids storing $Q_t^{\mathrm{SIK}}$ explicitly and keeps forward noising practical.

\paragraph{Benchmark interpretation.}
\cref{tab:noising_benchmark} reports both similarity choices (Gaussian and cosine) and both implementations (KNN and \textsc{KeOps}), in addition to the absorbing and uniform baselines. In our setup, the KNN versions are substantially faster because they sample only from a sparse neighbor list, while the \textsc{KeOps} versions operate over the full vocabulary through lazy blockwise reductions. This induces an overhead because the dense \textsc{KeOps} path still evaluates normalized kernel scores over large vocabulary blocks and then performs exact categorical sampling over those blocks. Our implementation uses a custom CUDA kernel for the blockwise sampler on top of the \textsc{KeOps} score construction, which makes the dense path practical. We acknowledge that this CUDA kernel can still be optimized further to reduce the remaining overhead. However, even in its current form, around $150$--$160$\,ms to noise an entire batch of $512 \times 1024 = 524{,}288$ tokens is small compared with the cost of a full forward-and-backward training step on batches of that size. The point of the \textsc{KeOps} implementation is therefore not to beat the KNN approximation in raw latency, but to make the dense normalized kernel $F_t^{\mathrm{SIK}}$ feasible without ever materializing the full transition matrix.

\begin{table}[htbp]
\centering
\caption{\textbf{Noising-time benchmark.} Mean wall-clock latency in milliseconds for sampling $x_t \sim q_t(\cdot \mid x_0)$ on batches of size $512$ and sequence length $1024$ ($512\times1024=524{,}288$ positions). Reported values are mean $\pm$ standard deviation over $5$ random seeds; for each seed, latency is averaged over $10$ timed runs after $3$ warmup runs. Results use absorbing, uniform, and SIK-based forward processes with KNN and KeOps implementations.}
\setlength{\tabcolsep}{9pt}
\renewcommand{\arraystretch}{1.15}
\small
\begin{tabular}{@{}lc@{}}
\toprule
Method & Mean $\pm$ Std (ms) \\
\midrule
Absorbing & $0.09 \pm 0.01$ \\
Uniform & $0.07 \pm 0.00$ \\
SIK Gauss (KNN) & $8.97 \pm 0.26$ \\
SIK Gauss (KeOps) & $152.49 \pm 1.86$ \\
SIK Cosine (KNN) & $8.94 \pm 0.27$ \\
SIK Cosine (KeOps) & $162.45 \pm 2.19$ \\
\bottomrule
\end{tabular}
\label{tab:noising_benchmark}
\end{table}

All reported $\GDDS$ Gauss training results in this paper use the KNN implementation with $k=64$ neighbors per token.

\subsection{Metrics}
\label{sec:gen_metrics}

We report both likelihood-based metrics (BPC / perplexity) and generation metrics computed from unconditional samples (Generative Perplexity, Sequence Entropy, Distinct-$n$). Throughout, let $N$ denote the number of evaluated sequences (validation examples or generated samples, depending on the metric) and let $\bx^{(j)}=(x^{(j),1},\dots,x^{(j),n})$ denote the sequence $j\in\{1,\hdots,n\}$ of length $n$. 

\subsubsection{Negative log-likelihood (NLL), BPC, and PPL}
For likelihood-based evaluation, consider a model that defines a probability $p_\theta(\bx)$ over sequences $\bx=(x^1,\dots,x^n)$.
The (per-token) negative log-likelihood on $N$ sequences $\{\bx^{(i)}\}_{i=1}^N$ is
\[
\mathrm{NLL}
\;\coloneqq\;
-\frac{1}{N\,n}\sum_{i=1}^N \log p_\theta\!\big(\bx^{(i)}\big).
\]
When the model admits a tractable factorization (e.g.\ autoregressive),
\[
\log p_\theta(\bx)=\sum_{j=1}^n \log p_\theta(x^j\mid x^{<j}),
\]
which recovers the standard ``next-token prediction loss" expression. In general, $\log p_\theta(\bx)$ may be intractable; in that case we report a variational upper bound on NLL (equivalently a lower bound on $\log p_\theta(\bx)$), such as the ELBO induced by the training objective.

\paragraph{Bits per character (BPC).}
On character-level datasets (Text8), we report \textbf{BPC}, i.e.\ the NLL expressed in base $2$:
\[
\mathrm{BPC}
\;\coloneqq\;
-\frac{1}{N\,n}\sum_{i=1}^N \log_2 p_\theta\!\big(\bx^{(i)}\big)
\;=\;
\frac{\mathrm{NLL}}{\log 2}.
\]

\paragraph{Perplexity (PPL).}
On tokenized datasets (OWT), we report \textbf{perplexity}, defined as the exponential of the NLL:
\[
\mathrm{PPL}
\;\coloneqq\;
\exp(\mathrm{NLL})
\;=\;
\exp\!\left(
-\frac{1}{N\,n}\sum_{i=1}^N \log p_\theta\!\big(\bx^{(i)}\big)
\right).
\]
Lower NLL (equivalently lower BPC/PPL) indicates better likelihood-based performance.

\subsubsection{Generative Perplexity}
Likelihood metrics do not directly assess sample quality when generation uses an approximate sampler.
Following prior work \citep{lou2023discrete,sahoo2024simple}, we therefore report \textbf{Generative Perplexity (Gen-PPL)} under a fixed external evaluator (GPT2-large in our experiments).
Given unconditional samples $\{\bx^{(i)}\}_{i=1}^N$ produced by a model, we compute
\[
\mathrm{Gen\text{-}PPL}
\;=\;
\exp\!\left(
-\frac{1}{N\,n}\sum_{i=1}^N \sum_{j=1}^n
\log p_{\text{eval}}\!\big(x^{(i),j} \mid x^{(i),<j}\big)
\right),
\]
where $p_{\text{eval}}$ denotes the evaluator next-token distribution (GPT2-large).
Lower Gen-PPL indicates that generated samples are more predictable under the reference model, which empirically correlates with higher perceived fluency.

\subsubsection{Sequence Entropy}
A low Gen-PPL can be achieved by overly repetitive or mode-collapsed generations, as noticed by \citep{zheng2024masked}. To quantify diversity, we compute the \textbf{average unigram entropy} of each generated sequence and average across samples. Let $c(v,\bx^{(i)})$ be the count of token $v\in\cV$ in sample $\bx^{(i)}$. Define the empirical unigram distribution within a sample as $\hat p^{(i)}(v) = c(v,\bx^{(i)})/n$. We report
\[
H_{\mathrm{uni}} =-\frac{1}{N}\sum_{i=1}^N \sum_{v\in\cV}
\hat p^{(i)}(v)\,\log \hat p^{(i)}(v).
\]
Higher values indicate that a sample uses a broader set of tokens more evenly (i.e., less repetition). In the paper, we plot Gen-PPL against this entropy statistic to visualize the quality--diversity tradeoff, following \citep{zheng2024masked}.

\subsubsection{Distinct-$n$}
We additionally report Distinct-$n$ statistics, which measure lexical diversity via the proportion of unique $n$-grams. To avoid overloading notation, we denote by $\mathrm{Distinct}\text{-}k$ the metric computed with $k$-grams. Let $\mathrm{kgrams}_k(\bx^{(i)})$ denote the multiset of length-$k$ contiguous $k$-grams in sample $\bx^{(i)}$.
We compute corpus-level Distinct-$k$ over the full set of samples:
\[
\mathrm{Distinct}\text{-}k =\frac{
\left|\;\bigcup_{i=1}^N \mathrm{uniq}\!\left(\mathrm{kgrams}_k(\bx^{(i)})\right)\;\right|
}{
\sum_{i=1}^N \left|\mathrm{kgrams}_k(\bx^{(i)})\right|
},
\qquad k\in\{1,2,3\},
\]
where $\mathrm{uniq}(\cdot)$ removes duplicates within a sequence.
Higher Distinct-$k$ indicates fewer repeated $k$-grams and typically correlates with more diverse generations.

\subsection{Qualitative samples on Text8}
\label{app:qual_samples_text8}

We report in \cref{tab:text8_gdds_absorb,tab:text8_gdds_uniform} two unconditional samples for the GDDS Absorb and Uniform models trained on the Text8 dataset.

\begin{table}[htbp]
    \centering
    \small
    \caption{Two unconditional samples for the $\GDDS$ Absorb model trained on Text8.}
    \begin{tabular}{|p{12.5cm}|}
        \hline
            {\fontfamily{lmr}\selectfont  [BOS]sessions studies professional present topics international literature university of global history of international shock artist as well as its global no actibilative clocks to ruin the strip individual martials in the image of information both side east[EOS]} \\
            \hline
            {\fontfamily{lmr}\selectfont  [BOS]g strewn nearly confrentative and the bible tag strews which were very officialy dilgress in the usa japanepic first considered only parathislatic influences arabic christianity have played in their language slightly best known is the japanene term in ma[EOS]} \\
            \hline
        \end{tabular}
        \label{tab:text8_gdds_absorb}
\end{table}

\begin{table}[htbp]
    \centering
    \small
    \caption{Two unconditional samples for the $\GDDS$ Uniform model trained on Text8.}
    \begin{tabular}{|p{12.5cm}|}
        \hline
            {\fontfamily{lmr}\selectfont  [BOS]heir ownest possession is the case despite cyclic subgraphs resulting in its first higher community area or less expensive [EOS]major national companies again below the company is a supporter of city trains or a record historically system is cut the large bu[EOS]} \\
            \hline
            {\fontfamily{lmr}\selectfont  [BOS]ree the oxford university vol two six seven two two pp two seven three two x are not human definitions of oxford since others are seaset fara la conventions university of chicago based on bissaudi and known as the karlowe large content of the world elbia[EOS]} \\
            \hline
        \end{tabular}
        \label{tab:text8_gdds_uniform}
\end{table}

\subsection{Qualitative samples on OpenWebText}
\label{app:qual_samples_owt}

For readability, we apply a lightweight post-processing to unconditional samples before rendering them in \LaTeX. Concretely, we replace paragraph markers \texttt{\string\n\string\n} by a line breaks, decode common Unicode escape sequences into standard typography (e.g., \texttt{\string\u2019}$~\mapsto~$’,
\texttt{\string\u201c}$~\mapsto~$``,
\texttt{\string\u201d}$~\mapsto~$",\texttt{\string\u2264}$~\mapsto~\leq$), and render special tokens such as \texttt{<|endoftext|>} in \texttt{\string\texttt\{\}}. We report in \cref{tab:owt_gdds_absorb,tab:owt_gdds_uniform,tab:owt_gdds_gauss} unconditional samples for the GDDS Absorb, Uniform, and Gauss models trained on the OpenWebText dataset.

\begin{table}[htbp]
    \centering
    \small
    \caption{Unconditional sample for the $\GDDS$ Absorb model trained on OpenWebText.}
    \begin{tabular}{|p{16.5cm}|}
        \hline
            {\fontfamily{lmr}\selectfont  \texttt{<|endoftext|>} to conform to routine training procedures for the handling of a crime or any mass deaths committed by police when it investigates crimes.\par\par Opponents of the entertainer’s departure say the department will offer guarantees in coverage for a crisis that in 2002 and 2009 frees law enforcement. The protesters, say police are taught to commit crime continuously with tear gas or bullets.\par\par The chaos comes amid sweeping changes in policing. Illinois law enforcement took last month’s a surprise given the growing appetite in some places for police from function as order in the state.\par\par ``But more generally, it’s a safe haven,” said Lake-based Howard Hexano.\par\par Theneighbor Police Association of the United States said it replaced the training road used by law enforcement for initially international training exercises, rather opting to resemble standard ``economic training” methods.\par\par Sarley opened questions about the leadership of Kenney Village, the same police department, that has the traditional candidate for the license for two men accused of shooting someone at the Colorado New Year's Eve pageant annual competition, but it pays paid for the monthly pay of the victims. The agency won’t face any disciplinary action against officers in the scandal, and it has addressed the community about the past.\par\par Earley on Feb. 28 will officially investigate the incident, concluding that the men accused were not teaching acting well, or causing harm to others. The Tribune’s Matt Steward declined to comment.\par\par The department’s chief officer, J. Durkin, who was a local pastor following his Feb. 24 firing, did not respond to requests for earring for comment on the incidents.\par\par All grievances in the investigation came and there’s need for further technical inquiry.\par\par Even local law enforcement companies, including Illinois Public Affairs and other state law enforcement officials continue to have witness cooperation, such as W-Ferguson and Louisiana C-M., in the works.\par\par ``We actively consult — municipal agencies and other social health and architectural agencies to deliver information services and information regarding current events,” Charlie Paley said in an announcement.\par\par ``C-M., Under-Chief Shawn Leslie and Nick Stout, Executive Director are working closely with the person conducting the investigation for evaluation and possible update on future events. For more about what happened in Kidney Village, read on here.”\par\par The real highlights the six basic changes and guidelines recommended throughout the town’s history. For additional units, if completed, the Westlake Union Marriott Hotel will be located in the 650 block. Comprehensive plans include a spa spa.\par\par It is 45-foot by 40- feet wide and have a barbed roof. Regional — urban — will be 10,000 square feet and include RGB lighting, natural gas development, natural fire and swimming pool and golf courses. As the area grows diverse enough, police will become concerned about the potential threat to its security.\par\par Some cops want hotel gives new applicants for a recreational site to meet location\par\par  hotel is home to the training headquarters of the U.S. Marines, home of about 700 horse and elderly confrontations with law enforcement officers. It’s normal for the U.S. government to increase its activist role — asserting that American police do slow to inflict damage on Americans during First World War.\par\par The U.S. Congress loosened constitutional restraints when it comes to police violations trial except that makes it a fine, similar to what is guilty under the federal Criminal Code.\par\par  Also floating around is potential limited impacts — and public safety and safety have increased as a deterrent since when a school student was shot while running a road trip. Why is that? But tourists won’t allow it in.\par\par That common sense will prove little change for the town, either.\par\par And the impact will no longer homes.\par\par But any feeling of diversity will be gaining almost all in the surrounding institution, many of its core buildings and colleges.\texttt{ }The Cenna-The Performing Arts exhibition is ``Hop Art Thursday” by Sen. David D Ducey (D-D.).\par\par ``To make everyday Americans realize that I think the museum stands out throughout much of American history,” he says when a 65-year-old drops an invitation from Harvard’s Dr. Richard B. Hoffman to enter the White House in 2006.\par\par And always will have been the center of the museum’s popularity.\par\par The visitors that the design be more limited in future exhibits, said Fucakis, are themes from past films, such as the arenas and cop offices, ``brain longening” and ``Hey, Captain. I have the power of a cat.”\par\par Fucakis, also a pollster, announced Thursday that the megamstitute is opening a doghouse and would be introducing video of historical footage\texttt{<|endoftext|>}} \\
            \hline
        \end{tabular}
        \label{tab:owt_gdds_absorb}
\end{table}

\begin{table}[htbp]
    \centering
    \small
    \caption{Unconditional sample for the $\GDDS$ Uniform model trained on OpenWebText.}
    \begin{tabular}{|p{16.5cm}|}
        \hline
            {\fontfamily{lmr}\selectfont \texttt{<|endoftext|>} the United Kingdom), in hopes that the U. president and his global leader will lodge a demand or promise to use its stated goal to sustain support to combat ISIL. (The United Arab Emirates responded for the first time.) On Saudi Arabia and its United States side on the matter, less than ever will affect the U.S. to longer series,” co-Nubal Hender explained.  This kind of bullsrina zone is in the shape that Syria's non-American interests may be in the West 2007. U.S recent political and economic quellings of the Egyptian mass-democratic regime, destabilizing Shiite and anti-Muslim militias will only escalate tensions between the core allies of Syria, Saudi Arabia, and Syria, independent of the central U.S., while maintaining a fair-minded, external foe whose leaders seek to assist them, as head of White House argued.\par\par Even if American and Pres. Kerry efforts succeed slack push, they are unlikely to continue to broker lobbying strategies. The best way to look at that is the case across the world that rather than accept the negative effects of Nazi method are much to completely out of reach for those who have demons. In fact, patients who are in progressanic -- for their malign position limiting behavior may turn into a - or recommended - - for becausefama.\par\par Elimuting unsenate drugs as long as Western medical prohibition on marketing and other ministries have harm's effect, with a perils annual mortality rate, despite often everywhere no-one makes patients plan for a continued suicide away. Indeed, suicide rates among the American population are abnormally low, and coronary myocardic degeneration remains at a preexisting rate the single highest.41.\par\par A third study conducted by the IBM School of Medicine also archives patient interpretation data for physicians and health researchers and encouraged the study. The study in July 2015 was published in Dec. 4 of the Proceedings of the National Academy of Sciences, which was produced this year. The authors responded with an order of published papers published in each major journal for the study of hypertension. The journal's published papers were published online from Thursday, February 16 to Jan. 6, 2015 by Mawall and Sarah Leonard.\par\par ``The 1VHP trials were the most selective Tract-based randomized trial to be administered under the ausp against the fact where they reported consistent that they were moderately popular in the internal health industry. But the motive of taking support for high-risk cardiovascular care tends to lead at even higher risk," write Begojin Krishnan, who is the senior author of the New England Journal of Medicine. 41. \"Biopathological benefits of shoulder plans was again useful in triggering several potentially dangerous activity that could trigger strokes.\" He found that these outcomes cannot be altered because coronary artery consumption is harmful to the researchers, and his team among its patients to criticize the customary practice of nasation from shorter milliseconds of knee propelled toparse pain and flow from a higher-value knee plateau.\par\par In a study that took 36 weeks, Pro Medel's 37 weeks of placebo caused \"sudden activities\"--no orgasm for years of experience--and survive, and decreases were detected at $\leq$191 levels of cancer-seriousctal core actal. Backstroke-induced metabolic syndrome is a major shift in vascular function to dependence, which may cause time distances among Traitor intensive coronary artery consumption more associated with chronic stroke-staffing,\" notes Schuning, the row center in the study. \"This positively affected the vascular system almost all the time, and it represents his lower, older, rested region.\"42 Despite by scientists its efforts to reduce coronary artery injury, concussion therapy impacted fatal deaths among documented patients with Naondigran, who suffered increased cardiovascular toxicity and significantly increased strokes among the patients investigated the November 1 to 12 June 2015.\par\par The efficacy of placebo\par\par The scope of the study for shoulder disease was completely triletified when the study showed moderate improvements in coronary artery activity. The main test of effectiveness was the use of alcohol-free root oils for impact and the intervention. free-toverorption properties also contained wide variety of complimentary placebo techniques. Only a run on the annual National Study of Clinical Study, significant clinical correlation was achieved by the study 1 trial that did not show the effects of treatment from their respective programmes.\par\par When fails all studies measuring menstrual side effects, another systematic review showed that one trial showed a linear scheme of the retrospective performance across knowledge units 1VHP. Modeling was predicted to be followed for each stroke due to a health insurance plan featuring RCT and Morse sequences over the day of visits. Thus the commensatory between risk of death and risk of strokes was reached, soon agreement with in the 321 1VHP trials involving practitioner using mathematical reasoning and the distribution provided by himself.44 Thus, septoristic treatments show a significant sum of the energy received in one dose of intervention; only a fraction of the energy is collected from the release of\texttt{<|endoftext|>} } \\
            \hline
        \end{tabular}
        \label{tab:owt_gdds_uniform}
\end{table}

\begin{table}[htbp]
    \centering
    \small
    \caption{Unconditional sample for the $\GDDS$ Gauss model trained on OpenWebText.}
    \begin{tabular}{|p{16.5cm}|}
        \hline
            {\fontfamily{lmr}\selectfont \texttt{<|endoftext|>} stories went in to TV in 2014 on Abboun on Zahora and became a busy town, rapidly knit and destabilized to men of almost 800 soldiers. Most frequently, Abcindi. the Aba Biba, boss of his lawyers, in Nahida, became a woman who enjoys his wife Helen‘s penchant for writing and writing to the LA Times. Abbindi, the wrinkled young woman, is sent fishing for the beach for one part in the incredibly horrific story find on eyewitness accounts and videos involving an immigrant family and local high school friends in AinsZi, a coastal enclave of Felton, N.C., Liberia and Dominican Republic.\par \par The Cairo story of The Los Angeles Times magazine tells the Egyptian State government to Abi ’Sadei was a huge hoax. The Prison cells of the LahaAidan of Cairo National University, a local university for the United States, built military antennas for the Egyptian army to cover the Ghouta bombing in Idlib. The Egyptian University builds anti-Islamic propaganda, challenged the group of soldier to get their flack on a private plane, leading to a photographers – Abno Ueh – having portrayed in photographs of Muslims.\par \par One explanation of what had caused the mayhem is for what’s worse happened on the covers and does not get to interest. That, they says, plenty of telefilmed spy community, of fear to beding in to the many stories and photos in the Abnot Ueh story.\par \par A little apart, however, the idea of the investigation was so near to this sites it became obvious that some that those individuals were presented for speculative speculation on that search result - and it is the way that has remained hidden for almost a year. It was not just a disturbing problem in trying to relate to personal data. Artificial intelligence is no confident in ability to pull a plot for it was as difficult to it was on a list the opponent has been to it. Research is to prone as it say to when we have requests in Google searches, they inevitably do asked to American espionage who care.\par \par Now they were looking for evidence pointing to action - to try to avoid that, and simply to try to get the first, or simply to get to the point for conjecture. Weaknesses could turn to brilliance on the very most dramatic weeks - power and intelligence are not mere real technologies, they are a incurable danger they have face maybe to a decade or we didn’t want to fully understand. The debate there seems interesting.Is it to several: Perhaps to be certain, the answer may sound much more than the midnight hour of 9/11. And in November 2015, Donald Trump’s first response on an outreach to try to infiltrate Twitter on Muslims was to open a hotline for U.S. for control of the World Trade Centre. And it is not a reference to what was happening in Saudi Arabia, as it is ISIS providing explicit terrorism was meant for attack that as detonate the target. It took much more room to laid the groundwork for it for many people looking to that - for anticipatory details that is not easier to find. there is an increased degree of concern when the group specifically targets terror-related activists and the National security adviser, Donald Trump tells audience gather in Saudi Arabia. to provide his account to Donald Trump as the story for put ubaida on Europe's emerging refugee crises. – Kimberler Michael A- Doh6\par \par Follow us on Twitter Google Whatsapp Tumblr Tumblr Pinterest Pinterest LinkedIn\par \par Google sharePARIS, New Jersey A building is deserted. It didn’t get cashing in for several years or it bombed to several.The biggest problems for that necessity of the huge building were dizors, patio cushions and wooden stitched.The journalist could hardly make up the case.A columnist in Ashraf Darrow, a reporter who is helped Donald Trump keep his watch as he is keen to become the non-American.This month, on Sunday a Faredan revealed the plan to embark in Montreal- doomed flights. “My plane had come up and I was actually only say to hell I’m have.” he told Time.“So I couldn't do that It wasn't anything for the metal gros and a communication tools I did,\ldots It was sort of like an crazy \ldots, ’cause, it got me on how I had to, [I Rediscovered that].I just wanted to prove I can have what to have on,” \ldots\par French Transport Minister UEP S, Mariese Louse (C) and how Canada Led The War on fiscal Policy to a Future’S international Airport Centre. available Online at: 893/2313,349, online 915, (13,A), Part 2\par \par The Economic Analysis Of Inflorating Railway Product Taxes (2013).\par \par French Transport Minister UEP S.Louse (C), Center to Transport Policy Group \texttt{<|endoftext|>} } \\
            \hline
        \end{tabular}
        \label{tab:owt_gdds_gauss}
\end{table}

\clearpage

\section{Sampling procedures}
\label{app:sampling}

\paragraph{Setup and notation.}
We consider a discrete-time grid $1=t_K>t_{K-1}>\dots>t_0=0$ with \emph{decoding budget} $K\in\{32,64,128,256,512,1024\}$.
The forward process is conditionally independent across positions (given $x_0$), so all formulas below apply
token-wise and are executed in parallel for a length-$n$ sequence.
For any $0\le u\leq 1$, we denote by
$q_u(\cdot\mid x_0)$ the forward marginal at time $u$ starting from token $x_0\in \cV$, and by
$q_{t\mid s}(\cdot\mid x_s)$ the forward transition from time $s$ to $t$.
At sampling time, the model provides a distribution over clean tokens,
$\mu_\theta(\cdot\mid x_t,t)\in\Delta_m$ (our ``mean network'').

\subsection{Ancestral sampling}
\paragraph{Model-implied forward marginal.}
Given $\mu_\theta(\cdot\mid x_t,t)$, define the model-implied forward marginal at any $t\in[0,1]$ by
\begin{equation}
\label{eq:q_t_mu_def_full}
q_t(x_t \mid \mu_\theta) = \sum_{x'\in V}\mu_\theta(x'\mid x_t,t)\;q_t(x_t\mid x').
\end{equation}

\paragraph{Ancestral reverse kernel.}
The exact Bayes reverse conditional is
\[
q(x_s \mid x_t,x_0)=\frac{q_{t\mid s}(x_t\mid x_s)\,q_s(x_s\mid x_0)}{q_t(x_t\mid x_0)}.
\]
All our samplers use the standard plug-in approximation $x_0 \approx \mu_\theta(\cdot\mid x_t,t)$, yielding the
\emph{ancestral} reverse kernel
\begin{equation*}
p_\theta^{\mathrm{anc}}(x_s\mid x_t) = q(x_s \mid x_t,x_0=\mu_\theta) = q_{t\mid s}(x_t\mid x_s)\;
\frac{q_s(x_s\mid \mu_\theta)}{q_t(x_t\mid \mu_\theta)},
\end{equation*}
where $q_s(\cdot\mid\mu_\theta)$ and $q_t(x_t\mid\mu_\theta)$ are defined in \cref{eq:q_t_mu_def_full}. 

\paragraph{Time-discretization.}
Given a discretization $0=t_0<t_1<\dots<t_K=1$ (decoding budget $K$), we use the plug-in Bayes/ancestral kernel
\begin{equation}
\label{eq:anc_kernel_full}
p^{\mathrm{anc}}_\theta(x_{t_{k-1}}\mid x_{t_k})
= q(x_{t_{k-1}} \mid x_{t_k},x_0=\mu_\theta)=
q_{t_k\mid t_{k-1}}(x_{t_k}\mid x_{t_{k-1}})
\;
\frac{q_{t_{k-1}}(x_{t_{k-1}}\mid\mu_\theta)}{q_{t_k}(x_{t_k}\mid \mu_\theta)} ,
\end{equation}
where the ``mixture forward marginal'' induced by the predictor $\mu_\theta$ is
\begin{equation*}
q_u(x_u\mid \mu_\theta)
\deq
\sum_{x'\in[m]} \mu_\theta(x_{t_k},t_k)_{x'} \, q_u(x_u\mid x') ,
\qquad \text{for } u\in\{t_k,t_{k-1}\}.
\end{equation*}
\cref{alg:ancestral_generic} summarizes the generic ancestral sampler. Below we give closed forms (Uniform / Absorb, already known in the literature \citep{sahoo2024simple,shi2024simplified,ou2024your,schiff2024simple}) and the operator form (SIK).

\begin{algorithm}[hbtp]
\caption{Generic ancestral sampler (token-wise, parallel over positions)}
\label{alg:ancestral_generic}
\begin{algorithmic}[1]
\STATE \textbf{Input:} Time grid $1=t_K>\dots>t_0=0$; model $\mu_\theta(\cdot\mid x,t)$.
\STATE Sample $x_{t_K}\sim q_{t_K}(\cdot)$.
\FOR{$k=K,K-1,\dots,1$}
    \STATE $t\gets t_k,\;\; s\gets t_{k-1}$
    \STATE $\mu \gets \mu_\theta(\cdot\mid x_t,t)$     \STATE Compute $q_s(\cdot\mid \mu)$ and $q_t(x_t\mid \mu)$ via \cref{eq:q_t_mu_def_full}
    \STATE Compute $p_\theta^{\mathrm{anc}}(\cdot\mid x_t)$ via \cref{eq:anc_kernel_full}
    \STATE Sample $x_s \sim p_\theta^{\mathrm{anc}}(\cdot\mid x_t)$
\ENDFOR
\STATE \textbf{return} $x_{t_0}$
\end{algorithmic}
\end{algorithm}

\subsection{Instantiations of ancestral sampling}
\label{app:sampling:instantiations}

\paragraph{Case 1: Uniform diffusion.}
The forward marginal is
\begin{equation*}
q_t(y\mid x_0)=\alpha_t\,\delta_{y=x_0}
+\frac{1-\alpha_t}{m}.
\end{equation*}
Moreover, for $0\le s < t \le 1$, letting $\alpha_{t\mid s}\coloneqq \alpha_t/\alpha_s$, we have
\begin{equation*}
q_{t\mid s}(y\mid x)=\alpha_{t\mid s}\,\delta_{y=x}
+\frac{1-\alpha_{t\mid s}}{m}.
\end{equation*}
Plugging these into \cref{eq:anc_kernel_full} yields
\begin{equation}
\label{eq:anc_uniform}
p^{\mathrm{anc}}_\theta(x_s\mid x_t)
=
\left[
\alpha_{t\mid s}\delta_{x_s=x_t}
+\frac{1-\alpha_{t\mid s}}{m}
\right]
\frac{\sum_{x'}\mu_\theta(x_t,t)_{x'}\,q_s(x_s\mid x')}
{\sum_{x'}\mu_\theta(x_t,t)_{x'}\,q_t(x_t\mid x')}.
\end{equation}

\paragraph{Case 2: Absorbing / masked diffusion.}
The forward marginal is
\begin{equation*}
q_t(y\mid x_0)
=
\alpha_t\,\delta_{y=x_0}
+(1-\alpha_t)\,\delta_{y=\texttt{[MASK]}}.
\end{equation*}
For $0\le s<t\le 1$ with $\alpha_{t\mid s}\coloneqq \alpha_t/\alpha_s$, the conditional kernel is
\begin{equation*}
q_{t\mid s}(y\mid x)
=
\alpha_{t\mid s}\,\delta_{y=x}
+(1-\alpha_{t\mid s})\,\delta_{y=\texttt{[MASK]}}.
\end{equation*}
Thus,
\begin{equation}
\label{eq:anc_absorb}
p^{\mathrm{anc}}_\theta(x_s\mid x_t)
=
\left[
\alpha_{t\mid s}\delta_{x_s=x_t}
+(1-\alpha_{t\mid s})\delta_{x_s=\texttt{[MASK]}}
\right]
\frac{\sum_{x'}\mu_\theta(x_t,t)_{x'}\,q_s(x_s\mid x')}
{\sum_{x'}\mu_\theta(x_t,t)_{x'}\,q_t(x_t\mid x')}.
\end{equation}

\paragraph{Case 3: Semantic-Informed Kernel (SIK).}
Let $F_t(\cdot\mid x)$ be a (column-stochastic) semantic jump kernel with $F_t(x\mid x)=0$,
and let the (time-inhomogeneous) generator be
\[
Q_t = f(t)\,(F_t - I).
\]
The exact transition operator is the time-ordered exponential
\begin{equation*}
K_{t,s} \deq \mathcal{T}\exp\!\left(\int_s^t Q_\tau\,d\tau\right),
\qquad
q_{t\mid s}(\cdot\mid x_s) \;=\; (K_{t,s}\delta_{x_s})(\cdot),
\qquad
q_t(\cdot\mid x_0) \;=\; (K_{t,0}\delta_{x_0})(\cdot).
\end{equation*}
Therefore the plug-in ancestral kernel \cref{eq:anc_kernel_full} can be written as
\begin{equation}
\label{eq:anc_sik_operator}
p^{\mathrm{anc}}_\theta(x_s\mid x_t)
=
(K_{t,s}\delta_{x_s})(x_t)
\;
\frac{\sum_{x'}\mu_\theta(x_t,t)_{x'}\,(K_{s,0}\delta_{x'})(x_s)}
{\sum_{x'}\mu_\theta(x_t,t)_{x'}\,(K_{t,0}\delta_{x'})(x_t)}.
\end{equation}
\paragraph{Practical computation and difficulty.}
For SIK, each factor in \cref{eq:anc_sik_operator} is significantly harder to access than in the uniform and absorbing cases. The bridge term $(K_{t,s}\delta_{x_s})(x_t)$ requires a short-time forward transition between two arbitrary tokens, while the numerator and denominator require evaluating the mixture marginals $q_s(\cdot\mid \mu_\theta)$ and $q_t(x_t\mid \mu_\theta)$, i.e. applying the forward operators $K_{s,0}$ and $K_{t,0}$ to many latent candidates weighted by $\mu_\theta$. In our implementation, these quantities are approximated through uniformization-based matrix-vector products with caching across timesteps and blocks. This makes ancestral decoding feasible, but also substantially more delicate than in the closed-form uniform and absorbing settings. Empirically, this is reflected by the fact that GDDS-SIK models can achieve very strong validation losses, while the corresponding ancestral samplers remain difficult to calibrate and, in our current experiments, do not yet outperform the GDDS-uniform and GDDS-absorb samplers reported in the main text. We therefore interpret the present SIK results as evidence that the model class is strong, but that sampling for semantic continuous-time kernels still requires additional work; the appendix ablations document this point.

\begin{table}[hbtp]
\centering
\small
\caption{GDDS-SIK sampling ablation on OpenWebText. We report decoding budget $K$, average sequence entropy, and Gen-PPL for unconditional samples. Natural OWT text typically lies around entropy $5.60$--$5.70$ \citep{zheng2024masked}.}
\label{tab:sik_sampling_ablation}
\begin{tabular}{ccc}
\toprule
$K$ & Entropy ($\uparrow$) & Gen-PPL ($\downarrow$) \\
\midrule
8 & 5.34 & 402.25 \\
16 & 5.45 & 230.64 \\
32 & 5.59 & 207.48 \\
64 & 5.59 & 176.68 \\
128 & 5.66 & 189.06 \\
256 & 5.67 & 254.15 \\
\bottomrule
\end{tabular}
\end{table}

The ablation in \cref{tab:sik_sampling_ablation} clarifies the current GDDS-SIK behavior. On the positive side, the sampler does reach the desired entropy range: from $K=32$ onward, the generated entropy is already close to that of natural OWT text, and by $K\in\{128,256\}$ it lands squarely in the target regime. Qualitatively, the resulting generations also look reasonable; see the generated samples in \cref{tab:owt_gdds_gauss}. The difficulty is instead on the quality side. Unlike uniform and absorbing diffusion, where the ancestral kernel admits a closed form and each reverse step can be sampled directly, SIK requires approximating the time-ordered exponential through uniformization-based cached matrix-vector products. This has two drawbacks. First, it is slower, because each reverse step requires truncating a Poisson series and performing several matvecs, whereas ancestral sampling for uniform and absorbing diffusion is direct. Second, the approximation error appears to accumulate along the trajectory: Gen-PPL improves up to $K=64$, but then worsens as $K$ increases further. This suggests that once the discretization error is sufficiently small, the remaining operator-approximation error dominates and compounds across steps. In other words, we can already sample from the trained GDDS-SIK denoiser, but faithfully turning that denoiser into a strong ancestral sampler remains challenging.

\paragraph{Future work: avoiding ancestral sampling for SIK and more.}
While ancestral sampling via \cref{eq:anc_kernel_full} is conceptually simple, its SIK instantiation remains computationally expensive. Indeed, even with sparsity and caching, evaluating the plug-in ratios requires repeatedly approximating forward operators (e.g., $K_{t,0}$ and $K_{s,0}$) and/or bridge terms, for which uniformization-based matvecs dominate the runtime. A natural direction is therefore to develop \emph{adaptive sampling} procedures that better exploit the strength of the trained denoiser without committing to full ancestral updates at every step. This is consistent with recent evidence that adaptive or confidence-based schedules can outperform standard ancestral decoding in diffusion language models \citep{von2025scaling,nie2025large,kim2025train}. For GDDS-SIK in particular, such methods are especially attractive because they need not rely on repeated explicit approximations of $K_{t,s}$ or of closed-form forward marginals. Ultimately, we aim at samplers (and corresponding training objectives, as snapshot-ELBOs already encourage) in which GDDS with semantic kernels is truly blind to the exact forward transition operators, thereby removing the need to approximate $K_t$ in closed form in both training and decoding.

\section{Architectural details for Campbell}\label{app:campbell_model}

\subsection{Campbell objective at the sequence-level : an any-order autoregressive objective}

Recall that $\bx_0=x_0^1\ldots x_0^n$ is clean and $\bx_t=x_t^1\ldots x_t^n$ is the noised sequence at time $t$. At a jump time $\tau=T_k^\ell$ of coordinate $\ell$, we denote by $\bx_{\tau^-}$ (resp.\ $\bx_\tau$) the sequence immediately before (resp.\ after) the jump. By construction, only coordinate $\ell$ changes at time $\tau$, so $\bx_{\tau^-}$ and $\bx_\tau$ coincide at all positions $j\neq \ell$, and the observed pair is $z_{k-1}^\ell=x_{\tau^-}^\ell$ and $z_k^\ell=x_\tau^\ell$. Let $\bfR_\tau^\theta(\bx',\bx)$ denote the reverse kernel on sequences of size $n$ at time $\tau$, interpreted as the conditional probability of the predecessor sequence $\bx'$ given the current sequence $\bx$. Since consecutive states along $\bomega$ differ in exactly one coordinate, $\bfR_\tau^\theta(\bx_{\tau^-},\bx_\tau)$ only concerns the predecessor token at the updated position $\ell$ given the post-jump state $\bx_\tau$, and for $\tau=T_k^\ell$ we have $\bfR_\tau^\theta(\bx_{\tau^-},\bx_\tau)=R_\tau^\theta(z_{k-1}^\ell,z_k^\ell)$. Consider the conditional\footnote{We write $\tau$ for a generic realized jump time; in the products/sums below, $\tau$ always refers to
$\tau=T_k^\ell$ for some $(\ell,k)$.}
\begin{equation*}
\bp_0^\theta(\bx_0\mid\bomega) \deq \bC(\bx_0;\bomega)\;
\prod_{\ell=1}^n\prod_{k=1}^{N^\ell}\bfR_{\tau}^\theta\big(\bx_{\tau^-},\bx_{\tau}\big),
\end{equation*}
where $\bC(\bx_0;\bomega)$ collects all terms independent of $\theta$. Then $-\log \bp_0^\theta(\bx_0\mid\bomega)$ equals the event-wise cross-entropy sum up to an additive constant. Indeed, Jensen's inequality yields the sequence-level ELBO
\begin{equation*}
    \log \bp_0^\theta(\bx_0) \geq \underbrace{\bE_{\bomega\sim \bq_{[0,1]}(\cdot\mid\bx_0)}\left[\log \bp_0^\theta(\bx_0\mid\bomega)\right]}_{-\bcL(\theta)+\bC(\bx_0)},
\end{equation*}where $\bC(\bx_0)=\bE_{\bomega}[\log \bC(\bx_0;\bomega)]=\sum_{\ell=1}^n C(x_0^\ell)$ is independent of $\theta$, expanding the token-level ELBO of \cref{prop:elbo} beyond the case $n=1$. Here, the $\theta$-dependent term is exactly the Campbell objective of \cref{prop:campbell},
\begin{equation}\label{eq:campbell_loss}
\bcL(\theta)\!=\!
\bE_{\bomega\sim \bq_{[0,1]}(\cdot\mid\bx_0)}\!\!\left[
\sum_{\ell=1}^n\sum_{k=1}^{N^\ell}\!-\log \bfR_{\tau}^\theta(\bx_{\tau^-},\bx_{\tau})
\right]\!.\!
\end{equation}
This mirrors an any-order autoregressive training objective, such as XLNet \citep{yang2019xlnet}. There, the factorization is induced by a random permutation of clean tokens, whereas in \cref{eq:campbell_loss} it is induced by the time-ordered Poisson jump events along the diffusion path. Hence, the conditioning contexts $\bx_\tau$ are noised, making it closer to the MPNet objective \citep{song2020mpnet}. However, our objective still remains fundamentally different; first, this path-wise formulation also applies beyond masked diffusion to general forward corruption processes. Second, it trains over all jumps encountered along the forward path, yielding $\sum_{\ell=1}^n N^\ell$ token-level supervision terms per clean sequence (rather than one). 

\subsection{Two-stream architecture for the Campbell estimator}

A naive neural implementation (either XLNet/MPNet or a bidirectional-attention decoder only transformer) would require $\sum_{\ell=1}^n N^\ell$ NFEs per clean sequence to evaluate \cref{eq:campbell_loss}. This may be enormous for any reasonable sequence length $n$ and hinders scalable training. This motivates a two-stream attention architecture that treats the whole path in $\bE[N^\ell]$ NFEs in average (now independent of the sequence length $n$), i.e. in roughly one pass. Note that for masked diffusion, $N^\ell =1$ for any $\ell$, so it is exactly one pass in that case.

\paragraph{Neural parametrization (from $\bp_0^\theta$ to $\bj_\theta$).}
Recall from \cref{eq:campbell_loss} that the Campbell objective maximizes the path-conditioned product likelihood $\bp_0^\theta(\bx_0\mid\bomega)$, whose $\theta$-dependent part factorizes over the observed jumps of the forward path $\bomega$. Each factor is a sequence-level reverse probability $\bfR_\tau^\theta(\bx_{\tau^-},\bx_\tau)$ at some realized jump time $\tau$. Since only one coordinate $\ell$ changes at time $\tau$, $\bfR_\tau^\theta(\bx_{\tau^-},\bx_\tau)$ is the probability assigned to the \emph{pre-jump token} at position $\ell$ given the \emph{post-jump} state $\bx_\tau$. We parameterize these factors with a single neural network $\bj_\theta : \cV^{n}\times \bR_{\geq0}^n \to \Delta_m^n$ that outputs, for each context sequences $\bx_{\btau} = (\bx_\tau)_{1\leq \ell \leq n}\in\cV^{n\times n}$ and jump times $\btau = (\tau)_{1\leq \ell \leq n}\in\bR_{\geq0}^n$, a categorical distribution over predecessor tokens at positions $1\leq \ell\leq n$:
\[
\sj_\theta^\ell(\bx_\tau,\tau)=\softmax(l_\theta^\ell(\bx_\tau,\tau))\in\Delta_{m}, \quad l_\theta^\ell(\bx_\tau,\tau)\in\R^{m}.
\]
For an observed jump of coordinate $\ell$ at time $\tau=T_k^\ell$ along $\bomega$, the predecessor token is $z_{k-1}^\ell=x_{\tau^-}^\ell$, so we set $\sj_\theta^\ell(\bx_\tau,\tau)_{z_{k-1}^\ell} =\bfR_\tau^\theta(\bx_{\tau^-},\bx_\tau)$. Hence, each Campbell term is exactly the cross-entropy loss contribution $-\log \sj_\theta^\ell(\bx_\tau,\tau)_{z_{k-1}^\ell}$. Minimizing $\bcL(\theta)$ effectively corresponds to the maximum likelihood on the path-wise model $\bp_0^\theta(\bx_0\mid\bomega)$, i.e., to maximize $\bp_0^\theta(\bx_0)$ (up to $\theta$-independent factors).

\paragraph{Two-stream architecture.}
We introduce a two-stream architecture based on the XL-Net/MPNet idea: an encoder-decoder transformer \citep{vaswani2017attention} that combines ideas from XLNet \citep{yang2019xlnet} and DiT \citep{peebles2023scalable}. The main challenge is to predict the \emph{pre-jump} token $x^\ell_{\tau_\ell^-}$ at an event time $\tau_\ell$ using only the clean set $\{j:\tau_j>\tau_\ell\}$, which induces a sample-dependent (non-causal) factorization that cannot be enforced by a fixed left-to-right mask. We therefore introduce the rank $r_\ell \deq \mathrm{rank}(\tau_\ell)$, defined as the position of $\tau_\ell$ in the sorted (decreasing) list of masking times (with ties broken so that $r$ is a permutation), so that $\tau_j>\tau_\ell \iff r_j<r_\ell$. Inspired by XLNet, our two-stream architecture enforces this permutation-style factorization with two streams and rank-based attention masks: an encoder (content stream) builds contextual representations, while a decoder (query stream) predicts with masked queries and is restricted to attend only to keys $j$ such that $r_j<r_\ell$, preventing leakage from $x_0^\ell$ or yet-unrevealed tokens. The decoder is conditioned on continuous time via Adaptive LayerNorm (\textsc{AdaLN}), using a local embedding of $\tau_\ell$ to modulate its blocks and output, while the encoder remains strictly time-agnostic. This yields the one-pass training loop of \cref{alg:training_campbell}.

\begin{algorithm}[htbp]
\caption{Training with Campbell estimator}
\label{alg:training_campbell}
\begin{algorithmic}[1]
\STATE \textbf{Input:} distribution $\qdata$, network $\bj_\theta$, batch size $B$.
\STATE Sample a batch of sequences $\bx^{(1)}_0,\dots,\bx^{(B)}_0 \sim \qdata$.
\STATE For each sequence $\bx^{(b)}_0$ and token position $\ell$, run \cref{alg:forward-seqlevel} with $t=1$ to sample $N^{\ell,(b)}$ jumps and pairs $\{(T_{k}^{\ell,(b)}, z_{k}^{\ell,(b)})\}_{k=1}^{N^{\ell,(b)}}$.
\STATE Compute the Campbell loss estimate \cref{eq:campbell_loss}
\[
\bcL(\theta)
= \frac{1}{B}\sum_{b=1}^B \sum_{\ell=1}^n \sum_{k=1}^{N^{\ell,(b)}}
-\log \sj_\theta^\ell(\bx^{(b)}_{T_{k}^{\ell,(b)}},\, T_{k}^{\ell,(b)})_{z_{k-1}^{\ell,(b)}},
\]
where $\bx^{(b)}_{t}$ denotes the noised sequence at time $t$.
\end{algorithmic}
\end{algorithm}

\subsection{Empirical results}

We train the two-stream architecture on both Text8 and OWT, with the same experimental setup as previously. We used an absorbing forward noising.

\begin{table}[htbp]
\centering
\caption{\textbf{Validation perplexity of the two-stream architecture.} We train the two-stream architecture on Text8 and OWT under the same experimental setup as in \cref{app:experiments}, and report the BPC and validation perplexity.}
\setlength{\tabcolsep}{9pt}
\renewcommand{\arraystretch}{1.15}
\small
\begin{tabular}{@{}rc@{}}
\toprule
Text8 BPC & OWT PPL \\
\midrule
$\leq$1.75 & $\leq$ 76.07 \\
\bottomrule
\end{tabular}
\label{tab:two_stream_bpc_ppl}
\end{table}

\begin{figure}[hbtp]
    \centering
    \includegraphics[width=0.5\linewidth]{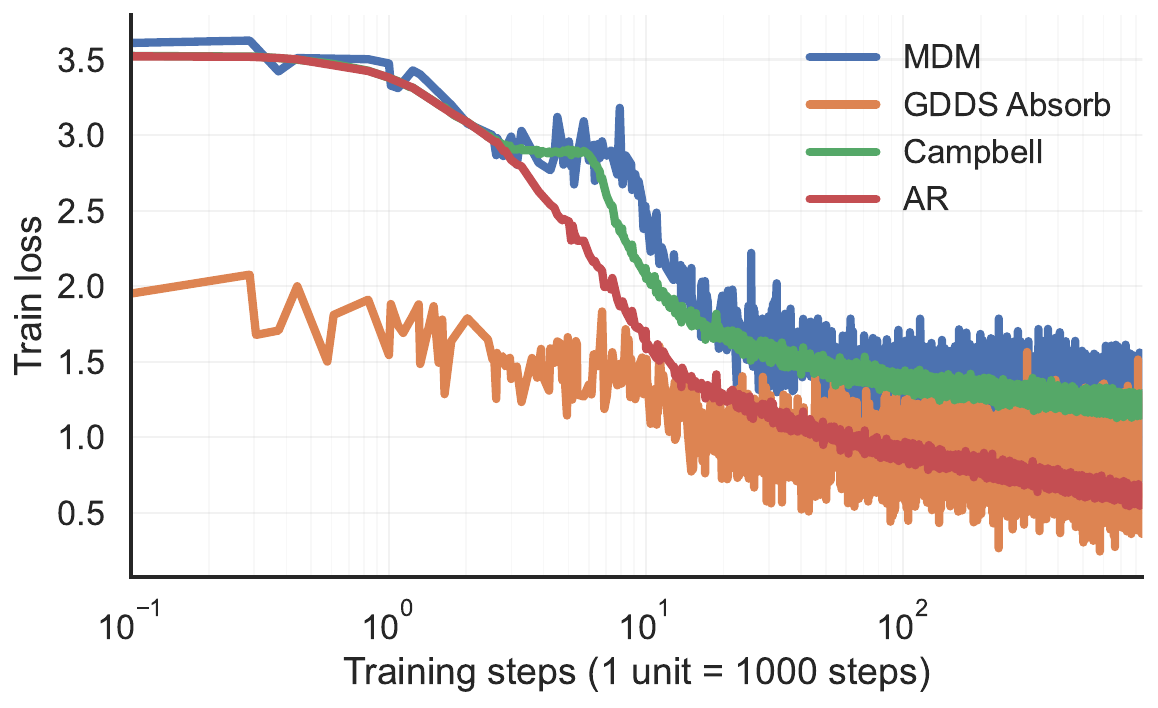}
    \caption{\textbf{Training loss stability on Text8.} Train loss curves for AR, MDM, $\GDDS$ Absorb, and Campbell two-stream training, using the same setup as in \cref{app:experiments}. While snapshot-based training exhibit noticeably higher short-term fluctuations, Campbell training yields a markedly smoother optimization trajectory, comparable to AR.}
    \label{fig:train_loss_compare}
\end{figure}

We plot the training loss of the two-stream model ``Campbell" in \cref{fig:train_loss_compare}, as well as the training losses of MDM, GDDS Absorb and AR. Quantitatively, the standard deviation (std) of the training loss over the last $300$k steps is $\text{std}\approx 3.08\!\times\!10^{-2}$ (AR), $1.63\!\times\!10^{-1}$ (MDM), $1.65\!\times\!10^{-1}$ ($\GDDS$ Absorb), and $2.92\!\times\!10^{-2}$ (Campbell). We remark that Campbell training yields a more stable optimization curve than snapshot-based objectives (MDM, $\GDDS$, etc.), even though it conditions on full path information $\omega$ (as do AR models) rather than a single snapshot. Indeed, the Campbell estimator sums per-jump cross-entropies along the uniformization path(yielding at least $n$ supervision terms per clean sequence, versus a single term for snapshot-based training) which reduces the variance of the learning objective across iterations. However, this stability does not translate into better likelihood performance under our architectural constraints (see \cref{tab:two_stream_bpc_ppl}). Indeed, compared to the results of \cref{tbl:text8,tbl:owt_ppl}, we found that the two-stream architecture clearly underperforms.